\documentclass[pdflatex,sn-mathphys-num]{sn-jnl}

\usepackage{graphicx}%
\usepackage{multirow}%
\usepackage{amsmath,amssymb,amsfonts}%
\usepackage{amsthm}%
\usepackage{mathrsfs}%
\usepackage[title]{appendix}%
\usepackage{xcolor}%
\usepackage{booktabs} 
\usepackage{makecell}
\usepackage{array}
\usepackage{textcomp}%
\usepackage{manyfoot}%
\usepackage{algpseudocode}
\usepackage{booktabs}%

\usepackage{algorithm}%
\usepackage{algorithmicx}%
\usepackage{algpseudocode}%
\usepackage{listings}%
\usepackage{soul}
\usepackage{xcolor}

\lstdefinestyle{jsonStyle}{
    language=JSON,
    basicstyle=\ttfamily\small,
    keywordstyle=\color{blue},
    stringstyle=\color{red},
    commentstyle=\color{gray},
    numbers=left,
    numberstyle=\tiny\color{gray},
    breaklines=true,
    frame=single,
    captionpos=b,
    showstringspaces=false
}

\theoremstyle{thmstyleone}%
\theoremstyle{thmstyletwo}%

\theoremstyle{thmstylethree}%

\begin{document}

\title[Article Title]{AgentChemist: A Multi-Agent Experimental Robotic Platform Integrating Chemical Perception and Precise Control}

\author[1]{\fnm{Xiangyi} \sur{Wei}}\email{52285901014@stu.ecnu.edu.cn}
\equalcont{These authors contributed equally to this work.}

\author[3]{\fnm{Fei} \sur{Wang}}\email{52290143075@stu.ecnu.edu.cn}
\equalcont{These authors contributed equally to this work.}

\author[2]{\fnm{Haotian} \sur{Zhang}}\email{10215304453@stu.ecnu.edu.cn}

\author[1]{\fnm{Xin} \sur{An}}\email{10225102428@stu.ecnu.edu.cn}

\author[3]{\fnm{Haitian} \sur{Zhu}}\email{52290143057@stu.ecnu.edu.cn}

\author*[3,4]{\fnm{Lianrui} \sur{Hu}}\email{lrhu@chem.ecnu.edu.cn}

\author*[1,4]{\fnm{Yang} \sur{Li}}\email{yli@cs.ecnu.edu.cn}

\author*[2]{\fnm{Changbo} \sur{Wang}}\email{cbwang@dase.ecnu.edu.cn}

\author*[3,4]{\fnm{Xiao} \sur{He}}\email{xiaohe@phy.ecnu.edu.cn}

\affil*[1]{\orgdiv{School of Computer Science and Technology}, 
\orgname{East China Normal University}, 
\orgaddress{\city{Shanghai}, \country{China}}}

\affil[2]{\orgdiv{School of Data Science and Engineering}, 
\orgname{East China Normal University}, 
\orgaddress{\city{Shanghai}, \country{China}}}

\affil[3]{\orgdiv{School of Chemistry and Molecular Engineering}, 
\orgname{East China Normal University}, 
\orgaddress{\city{Shanghai}, \country{China}}}

\affil[4]{\orgname{Shanghai Frontiers Science Center of Molecule Intelligent Syntheses}, 
\orgaddress{\city{Shanghai}, \country{China}}}



\abstract{Chemical laboratory automation has long been constrained by rigid workflows and poor adaptability to the long-tail distribution of experimental tasks. While most automated platforms perform well on a narrow set of standardized procedures, real laboratories involve diverse, infrequent, and evolving operations that fall outside predefined protocols. This mismatch prevents existing systems from generalizing to novel reaction conditions, uncommon instrument configurations, and unexpected procedural variations.
We present a multi-agent robotic platform designed to address this long-tail challenge through collaborative task decomposition, dynamic scheduling, and adaptive control. The system integrates chemical perception for real-time reaction monitoring with feedback-driven execution, enabling it to adjust actions based on evolving experimental states rather than fixed scripts. Validation via acid–base titration demonstrates autonomous progress tracking, adaptive dispensing control, and reliable end-to-end experiment execution. By improving generalization across diverse laboratory scenarios, this platform provides a practical pathway toward intelligent, flexible, and scalable laboratory automation.}




\maketitle

\section{Introduction}\label{intro}

Chemistry laboratory automation has been a well-established field for decades. Many existing systems~\citep{auto1, auto2, auto3, auto4, auto5, auto6, auto7, boiko2023autonomous, dai2024autonomous, zhu2024automated, rauschen2024universal, szymanski2023autonomous, merchant2023scaling, gao2025chemical} focus on automating standardized, repetitive tasks, and have made significant progress in improving efficiency and reproducibility. However, these systems often struggle to handle the diverse and evolving nature of real-world laboratory experiments~\citep{auto3, auto6, auto7, xu2025autonomous, song2025multiagent, tom2024self, m2024augmenting, mroz2025cross}. While previous works have optimized automation for well-defined protocols~\citep{auto1, auto2, leonov2024integrated, desai2025self, pijper2024addressing, gao2022autonomous}, they have not sufficiently addressed the challenge of automating tasks that follow a \textit{long-tail distribution}. In practice, a small number of standardized protocols are common, but a large variety of customized and exploratory tasks dominate daily research. Systems optimized for these “average” experiments face significant difficulty in adapting to the varied and unpredictable needs of research environments. Traditional systems are limited by three key failures that directly relate to the long-tail nature of laboratory work. First, they fail at {task generalization}, as they can only execute predefined protocols and are unable to compose new experimental workflows~\citep{auto6, auto7, m2024augmenting, zhang2025large}. Second, they suffer from a lack of {state perception}, meaning they do not monitor the progress of reactions in real-time and are instead constrained to rigid schedules~\citep{sun2025automated, pablo2025affordable}. Finally, their {failure in handling anomalies} makes it difficult to respond to unexpected experimental feedback, such as bubbles, color changes, or titration transitions~\citep{sinha2024real, lin2025visual}. Automation systems automate procedures, but not experiments.

\begin{figure}[h]
\centering
\includegraphics[width=\textwidth]{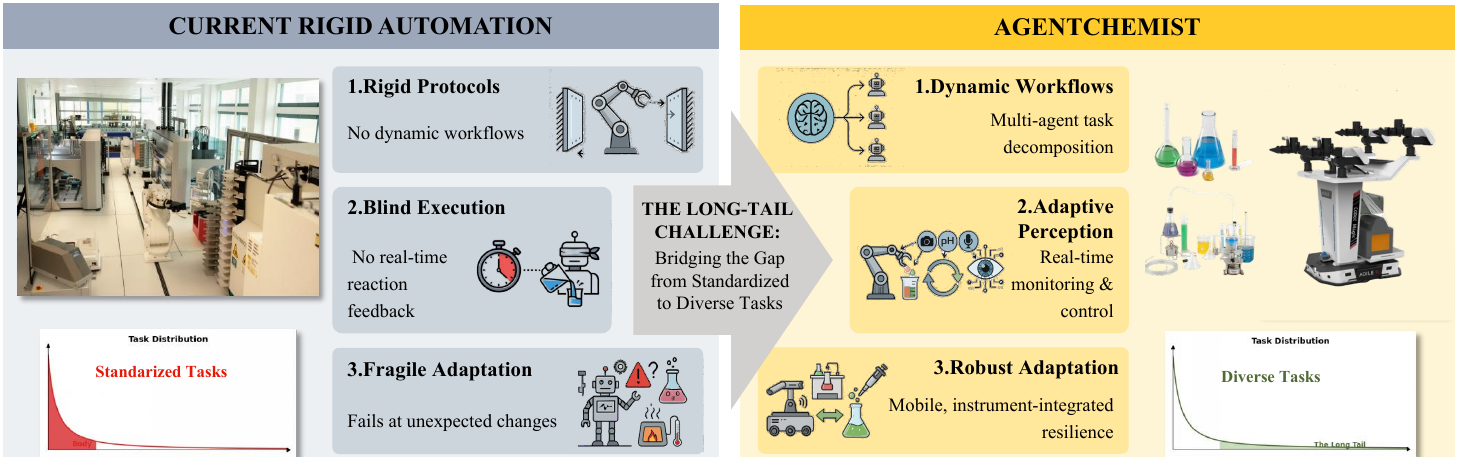}
\caption{The motivation for a multi-agent experimental robotic platform. Traditional rigid automation (left) is characterized by fixed protocols, blind execution, and fragile adaptation. In contrast, AgentChemist introduces dynamic workflows through multi-agent task decomposition, adaptive perception with real-time monitoring, and robust, instrument-integrated resilience to address long-tail challenges in the laboratory.
}
\label{motivation}
\end{figure}
The long-tail challenge in chemical experimental automation can be categorized into three main aspects. First, at the {task level}, current systems rely heavily on pre-scripted protocols, which limits their ability to adapt to new and diverse experimental workflows~\citep{anderson2006long}. Second, at the {chemical perception level}, most systems lack real-time monitoring capabilities, meaning they execute procedures based on fixed schedules rather than adjusting to the actual progress of reactions~\citep{xu2025autonomous, sun2025automated}. Finally, at the {laboratory environment level}, many systems are isolated from existing laboratory infrastructure~\citep{auto6, auto7}. They cannot easily integrate with pre-existing instruments or adapt to varying setups, which requires human intervention for tasks such as reagent replenishment and equipment reconfiguration. Furthermore, current systems often lack support for {human-robot interaction}, restricting their ability to collaborate with human researchers and adapt to dynamic, real-world environments~\citep{mroz2025cross, leong2025steering}. These limitations hinder the automation of true chemical experiments, where flexibility, adaptability, and collaboration are essential.

\begin{figure}[!t]
\centering
\includegraphics[width=\textwidth]{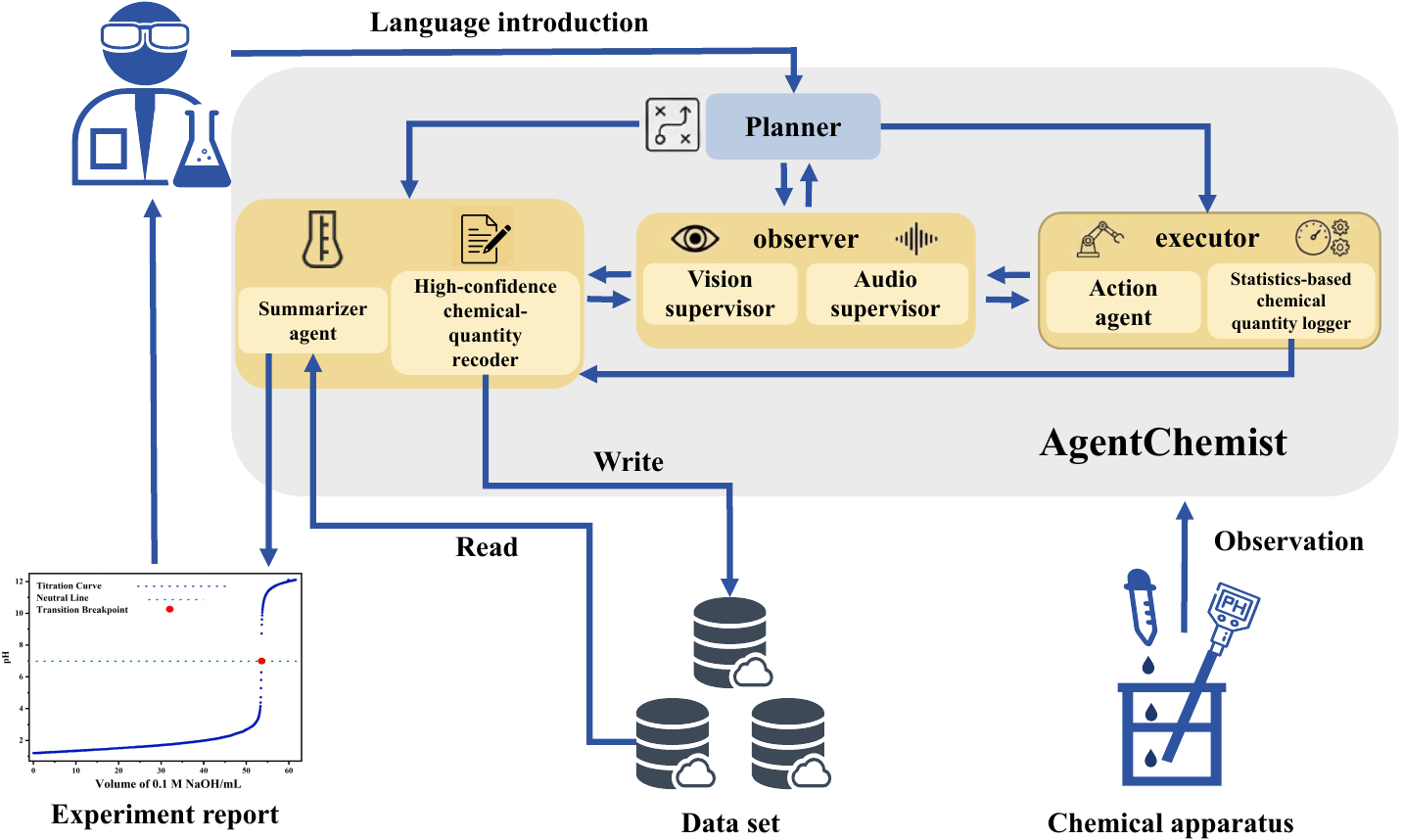}
\caption{AgentChemist Workflow: AgentChemist receives user instructions, plans experimental protocols, drives robots to perform specific experimental operations, records results in real time, and finally returns the experimental report to the user.
}
\label{schem}
\end{figure}

To address the long-tail challenges of chemical experimental automation, we propose the transition from traditional monolithic automation to a more robust multi-agent brain-integrated robotic execution approach~\citep{boiko2023autonomous, m2024augmenting}. As shown in Fig.~\ref{motivation}, the multi-agent plus robot approach provides a  framework for decomposing laboratory operations into specialized collaborative entities. Each agent can handle specific tasks, such as planning, perception, control, or logging, enabling the system to dynamically adapt to different tasks, continuously changing reactive states, and varying laboratory infrastructures~\citep{song2025multiagent,ruan2024automatic,park2023generative}. This design allows the system to tackle the long-tail nature of laboratory work, overcoming the task-level, chemical perception, and environmental challenges faced by traditional automation systems~\citep{gao2022autonomous}.
We introduce \textit{AgentChemist}, a Multi-Agent experimental robotic platform designed for intelligent, flexible chemical experimentation~\citep{openx2024embodied}.
AgentChemist is deployed on a general-purpose mobile chassis robot platform and integrates task-level planning, real-time experiment monitoring, precision control, chemical perception, and automatic report generation. 
AgentChemist treats the chemical experiment process as a Finite State Machine (FSM). Through communication among agents and the execution of subtasks, it drives this finite state machine from an initial state to an accepting state, thereby completing the experiment.

AgentChemist adopts a multi-agent collaborative framework as its core architecture. Fig.~\ref{schem} illustrates the interactive relationships between the user and AgentChemist, as well as among the internal agents within AgentChemist. Upon receiving a user's description of a chemical experiment, the Planner agent acquires environmental observations from the robot's sensors. Based on its interpretation of the instruction and environmental perception, it generates a finite state machine for the experiment, assigns subtasks to individual agents, and instantiates a statistics-based chemical quantity logger. The Vision supervisor and Audio supervisor function as observers within the framework. The Vision supervisor takes visual observations from the environment and feedback from other agents as input, overseeing the progression of the finite state machine and determining when to activate the executor. 
The Audio Supervisor takes contact-induced audio from the environment as input. Its primary functions are to provide estimates of chemical quantities derived from audio signals and to monitor the progress of the experiment.
The executor consists of the Action agent and the statistics-based chemical quantity logger, both of which are activated by the Vision supervisor and are responsible for driving the robot to perform actions. The Action Agent is a lightweight multimodal robot policy model responsible for driving the robot to execute  actions.
The Vision Supervisor oversees the evolution of the experiment's FSM. When a transition to a robot action is predicted to induce changes in chemical quantities, the Vision Supervisor activates a statistical chemical quantity recorder. It then assumes control of the robot from the action agent by executing finer-grained actions at a constant velocity, while providing chemical quantity estimates based on statistical analysis.
The High-Confidence Chemical Quantity Recorder receives statistical chemical quantity estimates, audio-derived chemical quantity estimates, state transitions occurring throughout the experiment, and any detected anomalies. It writes the chemical quantity data with the highest confidence level, along with detailed experimental records, into a dataset. Subsequently, the Summarizer Agent generates the final experiment report based on the data contained within this dataset.


AgentChemist overcomes the three key aspects of long-tail challenges in laboratory automation. At the {task level}, the multi-agent architecture allows the system to dynamically compose workflows and allocate tasks among specialized agents, adapting to diverse experimental protocols~\citep{boiko2023autonomous, zhang2025large}. At the {chemical perception level}, real-time sensing agents monitor parameters such as pH, color, volume, and temperature, adjusting dispensing and process parameters as needed to ensure precise control over experiments~\citep{xu2025autonomous, sun2025automated, pablo2025affordable, jiang2024artificial}. At the {environmental level}, AgentChemist is designed to integrate with a variety of laboratory instruments, reagents, and containers, thereby reducing reliance on specialized infrastructure~\citep{grosjean2025binding, swanson2025virtual}. As a result, it can operate flexibly across different environments, ensuring seamless operation in real-world laboratories without the need for custom-built tools or additional investments in specialized setups~\citep{leong2025steering}.

We validate the system through titration experiments and granular solid weighing experiments. The results show that AgentChemist can autonomously track experimental progress, adjust dispensing rates based on real-time data, and reliably complete multi-step procedures. These findings indicate that AgentChemist’s multi-agent approach mitigates long-tail challenges in chemical laboratory automation, improves chemical perception accuracy, and supports continuous operation for 8 hours without manual intervention.


\noindent This work makes the following contributions.
\begin{itemize}

    \item We propose \textit{AgentChemist}, a multi-agent experimental robotic platform deployed on a general-purpose mobile chassis robot. The platform supports task decomposition and agent-level coordination for planning, supervision, action execution, audio-assisted manipulation, and report generation. It enables end-to-end execution beyond fixed, pre-scripted protocols.

    \item We treat chemical experiments as FSMs and employ AgentChemist coupled with  robotic to drive the evolution of  FSMs. This approach addresses the long-tail distribution problems in chemical laboratory automation at three levels: task-level, chemical perception-level, and laboratory-environment-level.

    \item We demonstrate the AgentChemist on acid--base titration and granular solid weighing. AgentChemist tracks progress online, adapts dispensing rates from real-time signals, completes multi-step procedures, and generates reports. The system runs continuously for 8 hours without manual intervention.
\end{itemize}

Together, these contributions outline a pathway towards intelligent, flexible, and collaborative laboratory automation that is better suited to real-world research environments.

\section{Results}\label{sec2}





\subsection{Problem Definition}

AgentChemist defines a chemical experiment task as an agent collaboration process that can be formally described. This process includes four core stages: environment and instruction parsing, task decomposition, parallel execution, and result generation. Each experiment task is modeled as a five-tuple:
\[
\mathcal{E} = \{\mathcal{I}, \mathcal{S}, \mathcal{T}, \mathcal{O}, \mathcal{D}\}
\]

where:

\begin{itemize}
    \item \textbf{\(\mathcal{I}\)}: Set of natural language instructions. For example, \(i \in \mathcal{I}\) denotes ``Grab the rubber-tipped pipette, close the gripper tightly, and draw up the alkaline solution. Adjust the pH of the liquid in the beaker on the magnetic stirrer to 12.5''.
    
\item \textbf{\(\mathcal{S}\)}: Finite state machine representing the experimental process. Defined as \(\mathcal{S} = (Q, \Sigma, \delta, q_0, F)\), where:
\begin{itemize}
    \item \(Q\) is a finite set of states, including robot states \(s_r\) (e.g., joint positions, end-effector pose), environmental states \(s_e\) (e.g., instrument locations, task progress), and chemical states \(s_c\) (e.g., pH, temperature, color of solution, sound of liquid dripping).
    \item \(\Sigma\) is a finite set of input symbols representing agent actions and environmental events.
    \item \(\delta: Q \times \Sigma \rightarrow Q\) is the state transition function that defines how the system moves from one state to another based on inputs.
    \item \(q_0 \in Q\) is the initial state.
    \item \(F \subseteq Q\) is the set of final states representing task completion conditions.
\end{itemize}
The state machine formalism enables systematic tracking of experimental progression and ensures deterministic transition between operational phases.
\item \textbf{\(\mathcal{T}\)}:  Set of atomic sub-tasks of the agents. \(t \in \mathcal{T}\). 
    
    \item \textbf{\(\mathcal{O}\)}: Multimodal observation space. Comprises visual observations \(o_v\) (RGB images), auditory observations \(o_a\) (contact-based audio), and chemical observations \(o_c\) (sensor data like pH, conductivity).
    
    \item \textbf{\(\mathcal{D}\)}: Task-related data storage containing all critical experimental parameters recorded during the process.  Upon task completion, \(\mathcal{D}\) serves as the foundation for generating a structured experimental report.
\end{itemize}

\begin{figure}[!t]
\centering
\includegraphics[width=\textwidth]{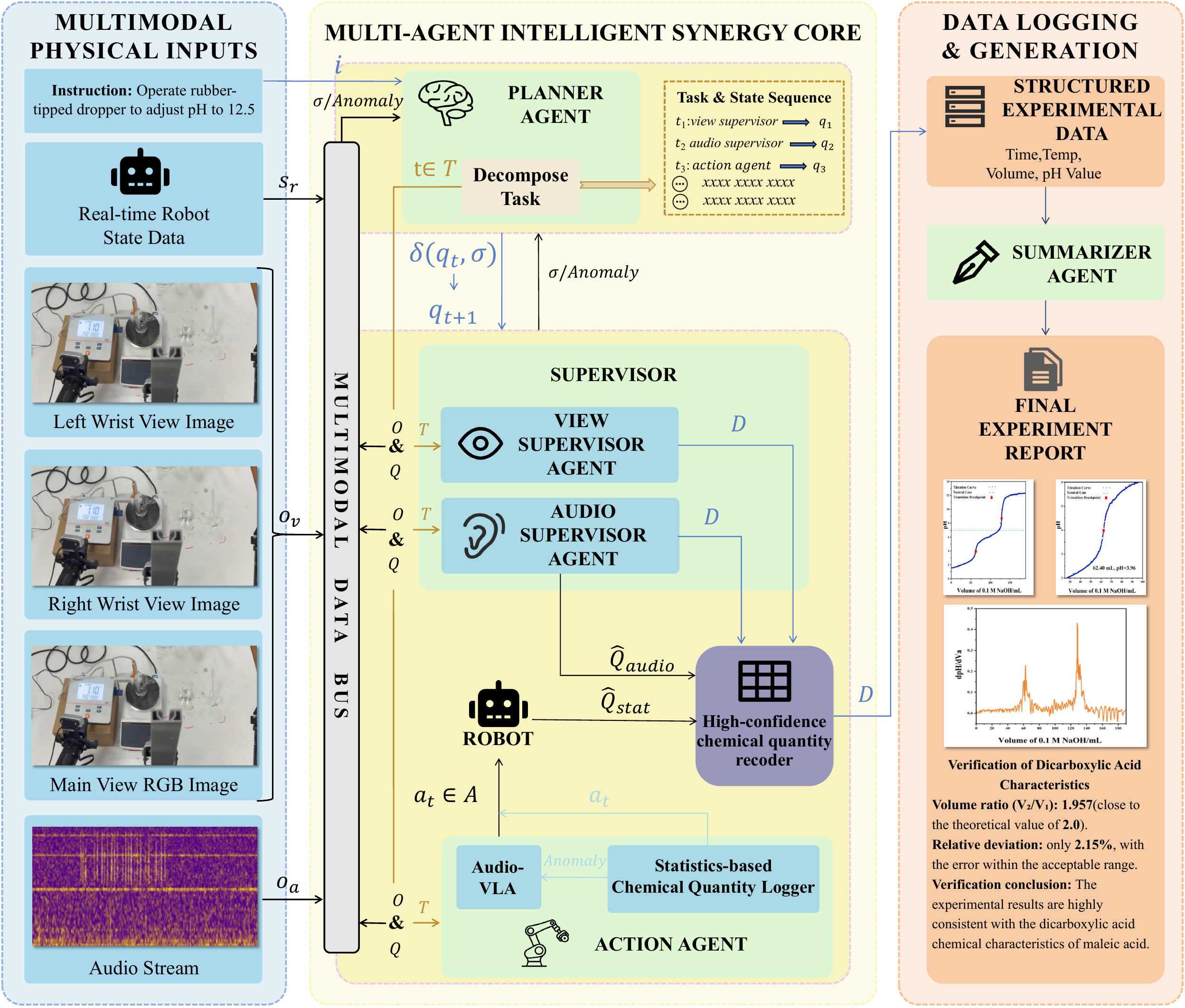}
\caption{AgentChemist workflow
}
\label{workflow}
\end{figure}

\subsection{AgentChemist workflow}

AgentChemist is a multi-level, modular multi-agent collaboration framework, which achieves a complete feedback loop from experimental instruction parsing to report generation through well-defined agent cooperation. AgentChemist consists of the planner agent, vision supervisor agent, Action agent, audio supervisor agent, summarizer agent, as well as the statistics-based chemical quantity logger and  high-confidence chemical quantity logging mechanism. Fig.~\ref{workflow} shows the workflow of AgentChemist.

Upon receiving a user's natural language description of a chemical experiment \(i \in \mathcal{I}\), AgentChemist initiates its workflow with the Planner Agent, which first parses the semantic meaning of instruction \(i\) and aligns it with real-time environmental observations \(\mathcal{O}\) acquired from robot sensors and the robot state \(s_r\) into a unified semantic space. Subsequently, the Planner generates a finite state machine \(\mathcal{S}\) that describes the experimental process.

Based on its understanding of the current experimental environment and task requirements, the Planner then decomposes the experiment into a set of atomic sub-tasks \(\mathcal{T}\), assigns them to the corresponding agents, and instantiates both the statistics-based chemical quantity logger and the high-confidence chemical quantity logging mechanism.

The Vision Supervisor Agent serves as the core observer and scheduler, continuously monitoring visual observations \(o_v\), chemical sensor data \(o_c\), and robot state \(s_r\). When the experiment requires robot actions, it schedules the Action Agent and the statistics-based chemical quantity logger, tracks the evolution of the state machine \(\mathcal{S}\), assesses experimental progress, and reports anomalies back to the Planner Agent.

The Action Agent, triggered by the Vision Supervisor Agent, is a policy model that drives the robot to execute actions \(a \in \mathcal{A}\) for performing chemical operations. When the Vision Supervisor Agent determines that the experiment has entered a phase involving changes in chemical quantities, it activates the statistics-based chemical quantity logger to execute robot actions and record statistically estimated chemical quantities. Concurrently, the Audio Supervisor Agent perceives physical events resulting from chemical quantity changes caused by robot motion through contact audio \(o_a\), providing audio-based estimates of chemical quantities and reporting potential anomalies.

Throughout the execution process, both the audio-based chemical quantity estimates and the statistics-based chemical quantity estimates, along with execution status and anomaly reports generated by all agents, are consolidated into the high-confidence chemical quantity logging mechanism instantiated by the Planner. This mechanism generates high-reliability structured data \(\mathcal{D}\) through a weighted fusion algorithm and updates the data storage in real time. When the state machine \(\mathcal{S}\) reaches the final state \(F\), the Summarizer Agent is triggered to automatically generate a structured experimental report based on the Planner's atomic sub-tasks, the complete state transition history, and the high-confidence data \(\mathcal{D}\), completing the full closed loop from instruction to result.

\subsection{Agents Design}
In AgentChemist, agents communicate through structured natural language messages, which are encapsulated in JSON tuples. Each message is formatted as follows:

\[
\{ \text{sender}, \text{receiver}, t, q_{\text{target}} \}
\]
where \( t \in \mathcal{T} \) represents the atomic sub-task to be executed, as defined in the task decomposition, and \( q_{\text{target}} \in Q \) is the expected state that the system should reach upon successful completion of the sub-task, corresponding to the target state of the finite state machine.

Upon receiving a message, the target agent parses the instruction \( t \) (sub-task) from the message and determines the necessary actions to perform.
It then executes the corresponding operation. Based on the operation performed, the agent drives the finite state machine \( \mathcal{S} \) toward the specified target state \( q_{\text{target}} \), ensuring the experiment progresses according to the predefined process. Through this structured communication mechanism, the agents in the system can work in coordination, maintaining alignment with the overall experimental process and correctly tracking the task progress through state transitions.

\subsubsection{Planner Agent}

The Planner Agent serves as the cornerstone of the AgentChemist system and is built upon the pre-trained large language model Qwen3-VL~\cite{qwen3-vl}. It is responsible for understanding unstructured natural language instructions \(i \in \mathcal{I}\), integrating them with visual observations of the experimental environment \(o_v \in \mathcal{O}\), and generating a structured executable workflow. 
The Planner Agent receives natural language instructions \(i \in \mathcal{I}\) from the user, which describe the experimental objectives, operations, and expected outputs. It also receives visual observations \(o_v \in \mathcal{O}\) of the experimental environment and the initial robot state \(s_r \in Q\) as inputs. During the semantic parsing phase, the Planner Agent extracts key information from the instructions and observations, including experimental entities, initial conditions, target conditions, operation intents, data recording requirements, and the expected output format. The Planner Agent then performs an environment verification by comparing the extracted requirements with the visual observations to confirm whether all necessary instruments and reagents are available. If any are missing, it generates feedback for the user.
After verification, the Planner Agent initializes the finite state machine (FSM) \(\mathcal{S} = (Q, \Sigma, \delta, q_0, F)\) that defines the entire experimental process. It then decomposes the experiment into a set of atomic subtasks \(\mathcal{T} = \{t_1, t_2, \ldots, t_n\}\), where each subtask represents the smallest operational unit that can be executed by an individual agent. The Planner Agent also instantiates the statistics-based chemical quantity logger and the high-confidence chemical quantity recorder. 
Once the processing is complete, the Planner Agent communicates with other agents according to the communication protocol, sending structured messages \(\{ \text{sender}, \text{receiver}, t, q_{\text{target}} \}\), where \(t \in \mathcal{T}\) is the atomic subtask to be executed, and \(q_{\text{target}} \in Q\) is the expected state after successful completion.

After processing, the Planner Agent distributes structured messages \(\{ \text{sender}, \text{receiver}, t, q_{\text{target}} \}\) to other agents according to the communication protocol, where \(t \in \mathcal{T}\) is the atomic subtask to be executed, and \(q_{\text{target}} \in Q\) is the expected state after successful completion. Specifically, the Planner Agent sends the complete task decomposition \(\mathcal{T}\) and the finite state machine \(\mathcal{S}\) to the Vision Supervisor Agent, making it responsible for overseeing the entire experimental process, tracking the evolution of the finite state machine, recording chemical quantities, receiving subtask completion reports from other agents, and driving state transitions based on event triggers. It sends subtasks requiring physical actions to the Action Agent, audio monitoring and audio-based chemical quantity recording tasks to the Audio Supervisor Agent, and report generation tasks to the Summarizer Agent. At the same time, the Planner Agent instantiates the statistics-based chemical quantity logger and the high-confidence chemical quantity recorder, configuring them for statistical chemical quantity estimation based on actuator displacement and fusion recording of heterogeneous data, respectively.
Upon receiving the messages, each agent parses the assigned subtask \(t\) and begins execution. Throughout the experiment, agents report completion events back to the Vision Supervisor Agent. The Vision Supervisor Agent uses these reports, along with \(\delta(q_t, \sigma) = q_{t+1}\), to drive the state machine transitions. This structured communication mechanism ensures that all agents work in coordination, maintaining consistency with the overall experimental process.

\subsubsection{Vision Supervisor Agent}

The Vision Supervisor Agent, as the core observer and scheduler within AgentChemist, is built upon the pre-trained Qwen3-VL multimodal large language model~\cite{qwen3-vl}. Its inputs include the task allocation table and finite state machine \(\mathcal{S} = (Q, \Sigma, \delta, q_0, F)\) received from the Planner Agent, as well as visual observations \(\mathcal{O}_v\) and chemical sensor data \(\mathcal{O}_c\) from the environment. The Vision Supervisor Agent is responsible for the continuous supervision of the experimental process, tracking state machine evolution, scheduling actuators, recording data, and managing anomalies.
Specifically, the Vision Supervisor Agent receives the complete task decomposition table \(\mathcal{T}\) and finite state machine \(\mathcal{S}\) from the Planner Agent, visual observations \(o_v \in \mathcal{O}_v\) (including wrist and front camera images), chemical sensor data \(o_c \in \mathcal{O}_c\) (such as pH, temperature, conductivity, etc.), as well as subtask completion reports and state updates from other agents. 
During the initialization phase, the Vision Supervisor Agent parses the task decomposition table \(\mathcal{T}\), identifies the monitoring subtasks assigned to it, and retrieves the initial state \(q_0\), accepting states \(F\), and the conditions for activating various actuators. In the execution phase, the Vision Supervisor Agent continuously supervises the experimental process, aligning the multimodal observations with the state definitions in the finite state machine \(\mathcal{S}\) to determine the current experimental state \(s_c \in Q\). It also receives subtask completion reports from other agents, verifies execution states and sensor status, and binds these events to the experimental state to drive the evolution of the state machine \(\mathcal{S}\).
The Vision Supervisor Agent infers the next state transition \(\delta(q_t, \sigma)\) and the target state \(q_{t+1} \in Q\) based on the reports from other agents, sends adjustment commands \(\sigma \in \Sigma\) to the Action Agent or the statistics-based chemical quantity logger, and continuously sends timestamps, chemical quantity measurements, current state \(q_t\), state transition events, anomaly event records, sent adjustment commands \(\sigma \in \Sigma\), and chemical quantity data obtained from the visual observations \(\mathcal{O}_v\) and chemical sensor data \(\mathcal{O}_c\) to the high-confidence chemical quantity recorder. This process continues the closed-loop control until \(s_c \in F\).

If anomalies or other issues are detected during monitoring or reported by other agents, the Vision Supervisor Agent records the anomaly information in the experimental data \(\mathcal{D}\) and sends the anomaly to the Planner Agent, which provides feedback to the user and waits for new planning instructions.

\subsubsection{Action Agent}

The Action Agent serves as the physical execution core in AgentChemist. It adopts a Vision-Language-Action (VLA) architecture and is embedded within a hierarchical control framework, responsible for transforming atomic subtasks \(t \in \mathcal{T}\) and control commands \(\sigma \in \Sigma\) into executable physical actions \(a \in \mathcal{A}\), thereby driving the state transitions of the finite state machine \(\mathcal{S}\). The Action Agent receives four types of heterogeneous inputs: natural language atomic subtasks \(t \in \mathcal{T}\) assigned by the Planner Agent, which specify the semantic intent of the operation to be executed; the current robot state \(s_r \in Q\), which is the robot's proprioceptive data in joint space; and multimodal observations \(o_v, o_a \in \mathcal{O}\), including RGB images from the main and wrist cameras and contact-based audio. Based on these multimodal inputs, the Action Agent \(\pi_{\text{VLA}}\) generates continuous motion control commands \(a_t = \pi_{\text{VLA}}(o_v, s_r, t, \sigma)\) end-to-end, where \(a_t \in \mathcal{A}\) represents the 7-DOF joint trajectory action. After execution, the result is abstracted as an input symbol \(\sigma \in \Sigma\), which is used to drive the state transition function \(q_{t+1} = \delta(q_t, \sigma)\). It is important to emphasize that the Action Agent does not directly modify the finite state machine \(\mathcal{S}\), but instead achieves the atomic subtasks specified by the Planner Agent and the symbolic state transitions defined by the Vision Supervisor Agent through physical execution. 

For each atomic subtask \(t \in \mathcal{T}\), the Action Agent first verifies whether the preconditions for execution are met based on the current robot state \(s_r \in Q\). Then, the VLA model generates continuous motion trajectories \(a_t\) using visual observations \(o_v \in \mathcal{O}\) and audio observations \(o_a \in \mathcal{O}\). The robot is driven to execute the action command via the lower-level interface, and after execution, the Action Agent reports the completion status to the Vision Supervisor Agent, triggering the corresponding state transition. The Action Agent adopts the Audio-VLA model architecture, consisting of four parts: the encoding layer, alignment layer, backbone network, and output layer. In the encoding layer, the main view image \(o_{\text{fv}} \in \mathcal{O}\) and text instruction stream \(t \in \mathcal{T}\) are input into the Florence-Base visual encoder, while wrist view images \(o_{\text{wv}} \in \mathcal{O}\) and other auxiliary views are processed by the Shared ViT with shared weights. Both are converted into visual token streams via patch splitting and linear embedding. The audio observations \(o_a \in \mathcal{O}\) are input into the Audio-CLIP audio encoder to generate audio token streams. The robot's proprioceptive data \(s_r \in Q\), time-step information, and initial action noise are concatenated repeatedly to form the raw control representation. In the alignment layer, the visual tokens, text instruction tokens, and audio tokens are aligned to a unified latent space through a shared linear projection layer. The raw control representation is then projected into control tokens through the ALOHA-specific input projection layer, achieving unified multimodal feature dimensions. In the inference layer, the multimodal token sequence is processed by a backbone network composed of 12 standard Transformer blocks. The global self-attention mechanism captures the deep spatiotemporal correlations between visual semantics, language instructions, audio events, and the robot's physical state \(s_r \in Q\). In the output layer, the fused control tokens are inverse-mapped by the ALOHA-specific output projection layer and reconstructed into a continuous sequence of joint trajectory points \(a_t \in \mathcal{A}\), directly driving the hardware to execute. After the action is executed, the Action Agent reports the completion status to the Vision Supervisor Agent, including execution success, timestamp, and key execution parameters, which serve as the input symbol \(\sigma \in \Sigma\) for state transition. This allows the Vision Supervisor Agent to drive the state machine evolution based on \(\delta(q_t, \sigma) = q_{t+1}\).

\subsubsection{Audio Supervisor Agent}

The Audio Supervisor Agent serves as the core auditory perception and event validation component in AgentChemist. It is built upon the pre-trained Qwen2-Audio-7B multimodal large language model~\cite{qwen2-audio} and is responsible for transforming contact-based audio observations \(o_a \in \mathcal{O}\) into quantifiable chemical event logs and state transition trigger signals. In the multi-agent framework, the Audio Supervisor Agent takes on the closed-loop role of perception, validation, and triggering, verifying event existence, providing audio-based chemical quantity estimates as closed-loop feedback, outputting state transition trigger signals to drive state machine evolution, and reporting anomalies when they exceed the system's autonomous handling capacity.

In the chemical laboratory environment, visual observations of transparent containers and colorless liquids \(o_v\) are often ambiguous due to lighting changes, surface reflections, and container refraction, making them unreliable as deterministic indicators of state transitions. In contrast, contact-based audio is inevitably associated with physical events, such as successful tool grabs or liquid droplets being added to the titration solution, which produce characteristic acoustic signals. These signals are invariant to lighting and have event determinism due to their underlying physical mechanisms, making contact audio an ideal modality for validating the existence of chemical quantity transfer events. Based on this physical prior, the Audio Supervisor Agent provides hard constraints for event existence to the high-confidence chemical quantity recording mechanism.

The Audio Supervisor Agent receives inputs including the task decomposition table \(\mathcal{T}\) and finite state machine \(\mathcal{S}\) from the Planner Agent, contact-based audio observations \(o_a \in \mathcal{O}\) sampled at 48 kHz with 16-bit quantization, and real-time state broadcasts \(s_c \in Q\) from the Vision Supervisor Agent. During the initialization phase, the Audio Supervisor Agent parses the task decomposition table \(\mathcal{T}\), identifies the audio monitoring subtasks assigned to it, and clarifies the types of audio events to listen for and the expected acoustic patterns during different experimental stages.

In the execution phase, the Audio Supervisor Agent continuously processes the audio stream and performs the following operations: first, it detects and segments events, accurately detecting and segmenting individual events in the continuous audio stream and estimating their start and end timestamps; next, it aligns and validates the state by using contextual information from the state representation \(s_c\) to logically validate the detected audio events. If the expected acoustic signal is detected during the execution of a subtask that is anticipated to generate that event, it is marked as a "expected event"; if the signal is detected without any corresponding operation instructions, it is marked as an "anomalous event." Following this, it performs chemical quantity estimation, estimating the chemical quantity change \(\hat{q}_{\text{event}}\) corresponding to each operation based on the acoustic features and accumulating the total chemical quantity \(\hat{Q}_{\text{total}} = \sum \hat{q}_{\text{event}}\), while providing a confidence score \(\gamma \in [0,1]\) for each estimate. Finally, it triggers state transitions, generating input symbols \(\sigma \in \Sigma\): expected events generate \(\sigma_{\text{event\_success}}\) to drive \(\delta(q_t, \sigma_{\text{event\_success}}) = q_{t+1}\), anomalous events generate \(\sigma_{\text{event\_abnormal}}\) to trigger the anomaly handling state, and long periods without detecting the expected event generate \(\sigma_{\text{event\_timeout}}\) to trigger a retry or reporting.

If anomalous events are detected, the Audio Supervisor Agent records the anomaly information in the experimental data \(\mathcal{D}\) and sends the anomaly to the Vision Supervisor Agent. Throughout the experiment execution, the Audio Supervisor Agent continuously sends timestamps, audio-based chemical quantity estimates \(\hat{Q}_{\text{audio}}\), confidence scores \(\gamma\), current states \(q_t\), detected event types and timestamps, anomalous event records, and generated state transition symbols \(\sigma \in \Sigma\) to the high-confidence chemical quantity recorder. When the state machine reaches an accepting state \(F \subseteq Q\), the Audio Supervisor Agent confirms the completion of chemical quantity recording and stops its work.

\subsubsection{Summarizer Agent}

The Summarizer Agent is built on the pre-trained Qwen3-VL model~\cite{qwen3-vl}and is responsible for integrating structured experimental data and task information into a clear and complete experimental report after the experiment is finished. The Summarizer Agent is activated only when the state machine \(\mathcal{S}\) reaches a terminal state \(F \subseteq Q\). It takes as input the structured experimental data output by the high-confidence chemical quantity recording mechanism, the state-transition history
\(\{(q_0, \sigma_1, q_1), (q_1, \sigma_2, q_2), \ldots, (q_{n-1}, \sigma_n, q_n)\}\), and the atomic subtask set \(\mathcal{T}\).

The Summarizer Agent first traverses the state-transition history and the experimental data \(\mathcal{D}\) to check whether any anomaly markers exist (e.g., sensor timeouts, operation failures, manual intervention). After completing anomaly auditing, the Summarizer Agent generates the experimental report by grounding on the subtask set \(\mathcal{T}\) issued by the Planner Agent and the structured experimental data \(\mathcal{D}\). Table~\ref{tab:report_modules} summarizes the structured knowledge extracted for report generation.

\begin{table}[h]
\centering
\caption{Structured Knowledge Extracted for Experimental Report Generation}
\label{tab:report_modules}
\begin{tabular}{|p{3cm}|p{6cm}|p{4cm}|}
\hline
\textbf{Module} & \textbf{Description} & \textbf{Data Source} \\
\hline
Experiment Metadata & Experiment time, task instruction \(i \in \mathcal{I}\), operator information, reagent batch numbers & \(\mathcal{I}\), system logs \\
\hline
Experimental Parameters & Initial conditions (volume, concentration, initial pH), target conditions (target pH, target volume) & \(\mathcal{D}\), \(\mathcal{T}\) \\
\hline
Operational Procedure & Atomic subtask sequence \(\mathcal{T}\) with execution timeline, agent assignments, dependency relationships & \(\mathcal{T}\), state-transition history \\
\hline
State Machine Evolution & Complete state transition history \(\{(q_0, \sigma_1, q_1), (q_1, \sigma_2, q_2), \ldots\}\), key state change points & state-transition history \\
\hline
Process Data & pH-time curve, volume-time curve, temperature-time curve, titration curve & \(\mathcal{D}\) \\
\hline
Chemical Quantity Records & Audio-based estimates \(\hat{Q}_{\text{audio}}\) with confidence \(\gamma_a\), statistical estimates \(\hat{Q}_{\text{stat}}\) with confidence \(\gamma_s\), fused high-confidence values \(\hat{Q}_{\text{fused}}\) & \(\mathcal{D}\), high-confidence recorder \\
\hline
Key Event Logs & Droplet detection events, state transition events, execution completion events & \(\mathcal{D}\), agent reports \\
\hline
Anomaly Records & Anomaly type, timestamp, system response (retry, adjustment, manual intervention), resolution status & state-transition history, \(\mathcal{D}\) \\
\hline
Experimental Results & Final pH value, total volume added, equivalence point analysis, pKa calculation results, relative deviation from theoretical values & \(\mathcal{D}\), task objectives \\
\hline
Visual Documentation & Key experimental snapshots (initial state, endpoint arrival, anomaly occurrence) & \(\mathcal{O}_v\) \\
\hline
Reproducibility Guide & Standardized description of experimental conditions, step-by-step operating procedures, parameter configurations & \(\mathcal{T}\), \(\mathcal{D}\) \\
\hline
\end{tabular}
\end{table}


\subsubsection{Statistics-based Chemical Quantity Logger}

In chemical experiments, certain operations directly impact the experimental results and require extremely high precision, such as liquid addition operations that demand fine-grained control. The Action Agent relies on human teleoperation demonstration data for imitation learning. However, due to the limitations of teleoperation, humans cannot guarantee the repeatability of high-precision operations, making it even more challenging to consistently reproduce such fine-grained manipulations during data collection. Consequently, the Action Agent is unable to infer high-precision control strategies from demonstration data. The introduction of the statistics-based chemical quantity logger compensates for this demonstration-execution precision gap.

When a state transition requires fine-grained control, the Supervisor Agent instantiates and activates the statistics-based chemical quantity logger. It can take over the low-level control of the Action Agent to execute precision operations that are inaccessible via teleoperation or demonstration, such as controlling the robot for liquid addition tasks. The statistics-based chemical quantity logger receives target control parameters from the Supervisor Agent, including the target chemical quantity \(s_c^{\text{target}}\), operation rate \(r_{\text{operation}}\), as well as the chemical state \(o_c \in \mathcal{O}\) and robot state \(s_r \in Q\) from the observation space, along with anomaly signals.

\begin{figure}[!t]
\centering
\includegraphics[width=\textwidth]{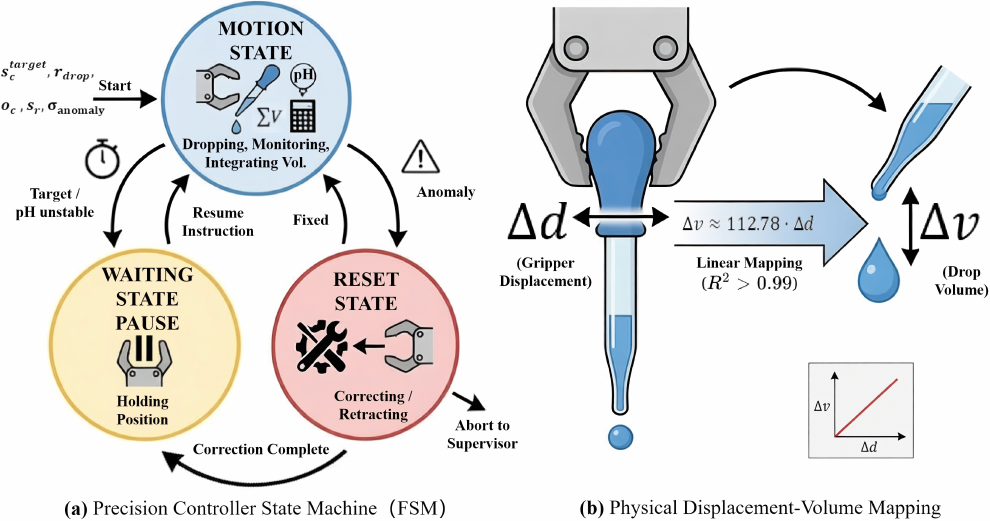}
\caption{State Machine of the Statistics-based Chemical Quantity Logger}
\label{contro}
\end{figure}

As shown in Fig.~\ref{contro} (a), the statistics-based chemical quantity logger operates an independent three-state finite state machine as its core control template, consisting of the motion state, waiting state, and reset state. In the motion state, the logger controls the actuator to move at a preset rate \(r_{\text{operation}}\), ensuring stable operation; it continuously monitors changes in the chemical state and statistically records the cumulative chemical quantity changes based on the \textbf{actuator displacement-chemical quantity mapping model}:
\begin{equation}
\Delta q = f(\Delta d) = k \cdot \Delta d
\end{equation}

where \(\Delta d\) represents the incremental displacement of the actuator, indicating the distance moved during the current control cycle; \(\Delta q\) is the chemical quantity change corresponding to a single operation; \(f\) is the actuator displacement-chemical quantity mapping function, established through offline calibration with a highly linear relationship; and \(k\) is the calibration constant, representing the chemical quantity change per unit displacement.

This mapping relationship is established through offline calibration, enabling low-cost, high-precision chemical quantity estimation without the need for external sensors such as flow meters. Based on this, the statistics-based chemical quantity logger accumulates the statistical chemical quantity estimate:
\begin{equation}
    \hat{Q}_{\text{stat}} = \sum \Delta q = \sum f(\Delta d)
\end{equation}

where \(\hat{Q}_{\text{stat}}\) is the cumulative estimated chemical quantity from the start of the experiment to the current time.

When the rate of chemical quantity change exceeds a threshold or approaches the target value \(|s_c - s_c^{\text{target}}| \leq \varepsilon\), the logger enters the waiting state, pausing actuator motion and maintaining the current position while waiting for subsequent instructions.

If an anomaly is detected (e.g., operation not executed properly, sensor timeout, operation timeout), a predefined error correction process is executed (such as retracting the actuator or repositioning), after which the system either returns to the motion state or reports to the Supervisor Agent. This state machine is initialized to the \textbf{motion state} when the statistics-based chemical quantity logger is activated, and during operation, it cyclically transitions between the three states based on chemical feedback and anomaly detection, until the task goal is achieved or the Supervisor Agent actively terminates the process.

The output of the statistics-based chemical quantity logger consists of two parts: the statistical chemical quantity estimate \(\hat{Q}_{\text{stat}}\) and the robot action command \(a \in \mathcal{A}\). The statistical chemical quantity estimate \(\hat{Q}_{\text{stat}}\) is sent to the high-confidence chemical quantity recording mechanism for data fusion and logging; the robot action command \(a \in \mathcal{A}\) directly drives the robot to execute specific physical operations, such as actuator movement and gripper opening/closing. These two outputs together form a complete closed loop from perception to execution.

\subsubsection{High-Confidence Quantity Logging Mechanism}

The high-confidence chemical quantity recording mechanism is designed to ensure the accuracy of recorded data. It is instantiated by the Planner Agent during the task initialization phase and runs continuously throughout the experiment. This mechanism is responsible for integrating chemical quantity observation data from multiple heterogeneous perception channels. Through confidence evaluation and weighted fusion strategies, it generates highly reliable structured experimental data.

The mechanism receives four types of inputs in real-time from three units: chemical sensor observations \(o_c\) (such as pH value, temperature, electrical conductivity, etc.) and their confidence \(\gamma_v\) along with timestamps, the current state machine status \(q_t \in Q\), and metadata such as the subtask context \(t \in \mathcal{T}\) from the vision Supervisor Agent; audio-based chemical quantity estimates \(\hat{Q}_{\text{audio}}\) and their confidence \(\gamma_a\) from the \textcolor{black}{audio supervisor agent}; and statistical base chemical quantity estimates \(\hat{Q}_{\text{stat}}\) and their confidence \(\gamma_s\) from the \textcolor{black}{statistics-based chemical quantity logger}.

For each observation input, the mechanism performs confidence evaluation based on the occurrence of physical events, generating a normalized confidence score \(\gamma \in [0,1]\). The confidence calculation not only depends on the signal quality of each perception modality but more importantly on {logical consistency verification with the system's execution state and abnormal events}. Specifically, the confidence of any chemical quantity observation is subject to the following three types of state constraints: first, if the current state machine status \(q_t \in Q\) indicates that the corresponding physical operation has not occurred (e.g., the system is in a waiting state or the \textcolor{black}{statistics-based chemical quantity logger} is not activated), but a perception channel still generates a non-zero chemical quantity record, the confidence of that channel is forced to zero; second, if the Supervisor Agent marks an operational failure (e.g., failed titration, sensor timeout) within the corresponding time window, the confidence of the relevant observation channel is reduced or set to zero based on the severity of the anomaly; finally, after passing the above logical verification, the signal confidence of each channel is normalized. The chemical sensor confidence \(\gamma_c\) is calculated based on the sensor's self-check status and signal noise level; the audio volume estimation confidence \(\gamma_a\) is output by the \textcolor{black}{audio supervisor Agent} based on acoustic feature clarity and event-state consistency; the statistical volume estimation confidence \(\gamma_s\) is computed based on the gripper calibration model residual and execution state.

Algorithm \ref{algo:hcqlm} illustrates the principles of the high-confidence chemical quantity recoder.
The high-confidence chemical quantity recoder employs a confidence-gated weighted fusion strategy to generate the final recorded value. Taking volume measurement as an example, it first eliminates observation channels with confidence below the threshold \(\theta = 0.6\). The confidence of the remaining channel set \(\mathcal{C}\) is then normalized to obtain the weights:
\begin{equation}
w_i = \frac{\gamma_i}{\sum_{j \in \mathcal{C}} \gamma_j}
\end{equation}
where \(w_i\) is the fusion weight for the \(i\)-th observation channel, \(\gamma_i\) is the confidence score of that channel, and \(\mathcal{C}\) is the set of remaining channels that pass the confidence threshold.

Next, the fused volume estimate is calculated:
\begin{equation}
\hat{V}_{\text{fused}} = \sum_{i \in \mathcal{C}} w_i \hat{V}_i
\end{equation}
where \(\hat{V}_{\text{fused}}\) is the fused volume estimate, \(\hat{V}_i\) is the volume estimate from the \(i\)-th observation channel, and \(w_i\) is its corresponding fusion weight.

The fused confidence is then computed as:
\begin{equation}
\gamma_{\text{fused}} = \frac{1}{|\mathcal{C}|} \sum_{i \in \mathcal{C}} \gamma_i
\end{equation}
where \(\gamma_{\text{fused}}\) is the global confidence score of the fused result, \(|\mathcal{C}|\) is the number of observation channels that pass the confidence threshold, and \(\gamma_i\) is the confidence score of the \(i\)-th channel.

For single-source chemical quantities such as pH value and temperature, the mechanism performs anomaly detection and smoothing filtering, and outputs a confidence label based on the sensor's self-check status.

The high-confidence chemical quantity data, after fusion calibration, are recorded in real-time as structured experimental data \(\mathcal{D}\). Each record contains a timestamp, chemical quantity name, fused estimate value, confidence score, raw values from each channel and their confidence, current state \(q_t\), and associated subtask \(t\). \(\mathcal{D}\) is persistently stored in a time-series database format and simultaneously pushed in real-time to the Summarizer Agent as the data foundation for report generation.

When the confidence of all observation channels falls below the threshold, the divergence between multi-source estimates exceeds the preset tolerance, or the chemical sensor readings are anomalous, the mechanism triggers a data quality anomaly flag and writes it into \(\mathcal{D}\). All raw observation data are fully retained to support full-link data traceability.
\begin{algorithm}
\caption{High-Confidence Chemical Quantity Recording Mechanism}\label{algo:hcqlm}
\begin{algorithmic}[1]

\Require Observation channels \(i \in \{\text{audio}, \text{stat}, \text{visual}, \text{sensor}\}\)
\Statex Observation values \(V_i\), raw confidence \(\gamma_i^{\text{raw}}\)
\Statex Current state \(q_t \in Q\), anomaly flags \(E \subseteq \Sigma\)

\Ensure Structured experimental data \(\mathcal{D}\)

\State \textbf{1. Confidence Evaluation (State-Event Consistency Check)}
\For{each observation channel \(i\)}
    \State \(\gamma_i \Leftarrow \gamma_i^{\text{raw}}\)
    \If{$q_t$ indicates no physical operation \(\wedge\) \(V_i \neq 0\)}
        \State \(\gamma_i \Leftarrow 0\)
    \ElsIf{$E$ contains execution anomalies within the current time window}
        \State \(\gamma_i \Leftarrow \gamma_i \times 0.1\)
    \EndIf
\EndFor

\State \textbf{2. Confidence-Gated Weighted Fusion}
\State \(\theta \Leftarrow 0.6\)
\State \(\mathcal{C} \Leftarrow \{ i \mid \gamma_i \geq \theta \}\)
\If{$|\mathcal{C}| > 0$}
    \For{each \(i \in \mathcal{C}\)}
        \State \(w_i \Leftarrow \dfrac{\gamma_i}{\sum_{j \in \mathcal{C}} \gamma_j}\)
    \EndFor
    \State \(\hat{V}_{\text{fused}} \Leftarrow \sum_{i \in \mathcal{C}} w_i V_i\)
    \State \(\gamma_{\text{fused}} \Leftarrow \dfrac{1}{|\mathcal{C}|} \sum_{i \in \mathcal{C}} \gamma_i\)
\Else
    \State \(\hat{V}_{\text{fused}} \Leftarrow \text{null}\)
    \State \(\gamma_{\text{fused}} \Leftarrow 0\)
    \State \textbf{Trigger} data quality anomaly ("All observation channels' confidence below threshold")
\EndIf

\State \textbf{3. Structured Data Recording}
\State \(\mathcal{D} \Leftarrow \{
    \text{timestamp}: \text{now}(),\;
    \text{volume}: \{\text{value}: \hat{V}_{\text{fused}},\; \text{confidence}: \gamma_{\text{fused}},\; \text{raw}: V_i\},\;
    \text{state}: q_t,\;
    \text{anomalies}: E
\}\)

\State \textbf{Return} \(\mathcal{D}\) to Summarizer Agent

\end{algorithmic}
\end{algorithm}

\subsection{An example of a closed-loop titration experiment driven by AgentChemist}

In a typical HCl-NaOH titration experiment, the user issues a natural language instruction to AgentChemist: ``Please grasp the rubber-headed pipette inside the beaker, use 0.1M sodium hydroxide solution to titrate the 0.1M concentrated hydrochloric acid with an initial pH of 1.06 placed on the magnetic stirrer, and terminate the titration when the pH reaches 12.5. You need to read the temperature and pH values from the pH meter, record the time and temperature at the beginning and end of the experiment, and record the pH value and the total volume of sodium hydroxide solution added after each drop. Do not execute actions when the pH is changing; continuously record the pH value and only execute actions when the pH stabilizes. Finally, based on the saved data, perform a detailed analysis to calculate the corresponding pKa value and generate an analytical report. The core requirement is to generate visual charts: Figure 1 is the titration curve, Figure 2 is its first derivative plot, Figure 3 is its second derivative plot, and Figure 4 is an enlarged view of the transition jump point. If any anomalies occur during the experiment or if the experimental equipment does not meet the requirements, you must inform me of the anomaly and the unmet condition.``

\begin{figure}[!t]
\centering
\includegraphics[width=\textwidth]{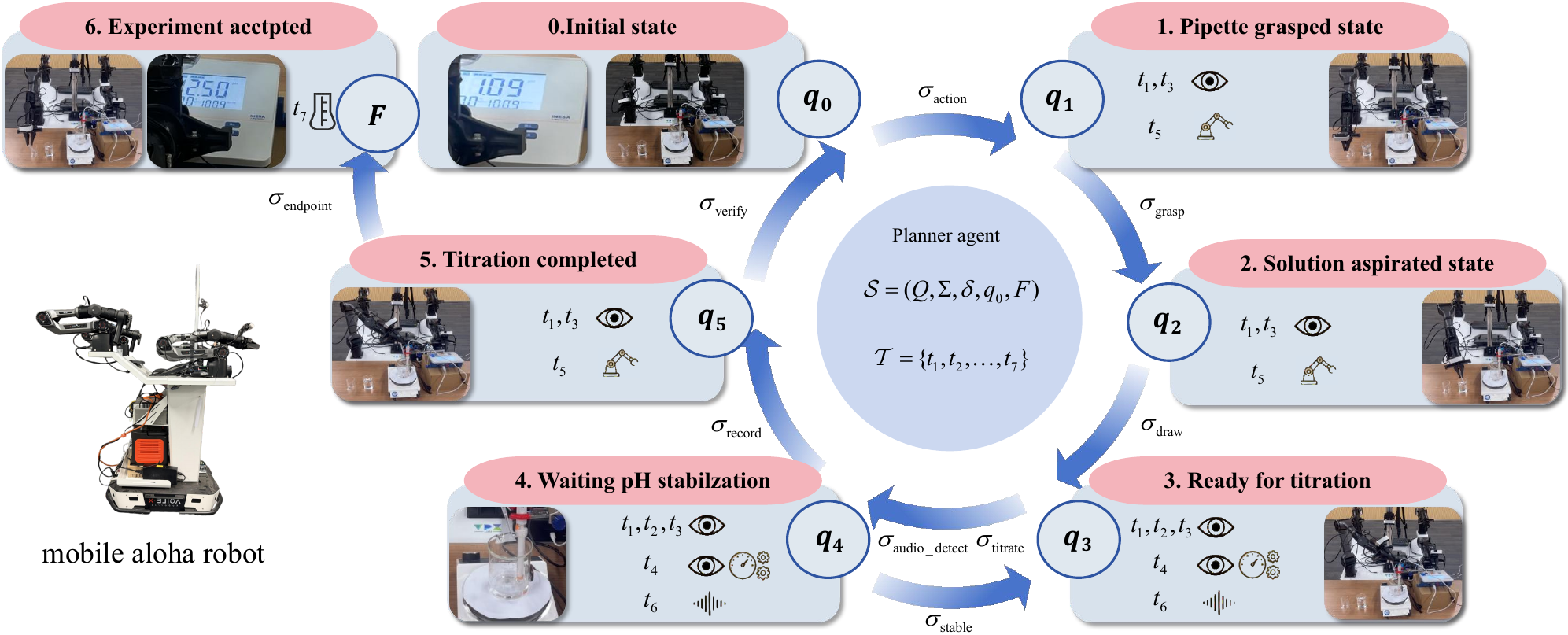}
\caption{The Process of AgentChemist Driving the Evolution of FSM \(\mathcal{S}\) in Acid-Base Titration Experiments}
\label{FSM}
\end{figure}

The Planner Agent serves as the cornerstone of AgentChemist, responsible for understanding unstructured natural language instructions \(i \in \mathcal{I}\), aligning them with visual observations of the experimental environment \(o_v \in \mathcal{O}\), and generating a structured executable workflow. Upon receiving the instruction \(i\), the Planner Agent first performs semantic parsing to extract key information: experimental entities including instruments such as the rubber-headed pipette, beaker, magnetic stirrer, and pH meter, as well as reagents including 0.1M sodium hydroxide solution and 0.1M concentrated hydrochloric acid; initial conditions of 25 mL concentrated hydrochloric acid with pH = 1.06; target condition of pH = 12.5; data recording requirements including recording pH and volume after each drop, and recording time and temperature at the beginning and end; output requirements of an analytical report containing four charts. As shown in Fig.~\ref{FSM}, after confirming through visual observations \(o_v \in \mathcal{O}\) that all required instruments and reagents are present on the experimental bench, the Planner Agent initializes the finite state machine \(\mathcal{S} = (Q, \Sigma, \delta, q_0, F)\). The state set \(Q\) includes the initial state \(q_0\), the state of having grasped the rubber-headed pipette \(q_1\), the state of having drawn the solution \(q_2\), the state of being ready for titration \(q_3\), the state of titration in progress \(q_4\), the state of waiting for pH stabilization \(q_5\), the state of titration completion \(q_6\), and the titration experiment accepting state \(F\). The input symbol set \(\Sigma\) includes symbols for driving the Action Agent \(\sigma_{\text{action}}\), grasp completion \(\sigma_{\text{grasp}}\), draw completion \(\sigma_{\text{draw}}\), driving the statistics-based chemical quantity logger to start titration \(\sigma_{\text{titrate}}\), the Audio Supervisor Agent detecting titration liquid \(\sigma_{\text{audio\_detect}}\), pH stabilization \(\sigma_{\text{stable}}\), data recording \(\sigma_{\text{record}}\), endpoint detection \(\sigma_{\text{endpoint}}\), and verification completion \(\sigma_{\text{verify}}\). Based on the finite state machine \(\mathcal{S}\), the Planner Agent decomposes the experiment into a set of atomic subtasks \(\mathcal{T} = \{t_1, t_2, \ldots, t_7\}\): \(t_1\) for monitoring the experimental process is assigned to the Vision Supervisor Agent. This subtask runs continuously, responsible for tracking the evolution of the state machine and aligning experimental events with state definitions to determine the current state \(s_c \in Q\). \(t_2\) for recording pH values is assigned to the Vision Supervisor Agent, corresponding to the input symbol \(\sigma_{\text{record}}\), and is expected to trigger data recording each time the pH stabilizes. \(t_3\) for driving the evolution of the state machine \(\mathcal{S}\) is assigned to the Vision Supervisor Agent, responsible for inferring the next state transition \(\delta(q_t, \sigma)\) based on reports from other agents and driving the system toward the target state. \(t_4\) for activating the statistics-based chemical quantity logger when the state machine transitions to the ready-for-titration state \(q_3\) is assigned to the Vision Supervisor Agent, enabling it to activate the statistics-based chemical quantity logger to perform titration operations upon detecting that the state has reached \(q_3\). \(t_5\) for driving the robot to complete the rubber-headed pipette liquid transfer operations is assigned to the Action Agent, including grasping the pipette corresponding to the input symbol \(\sigma_{\text{grasp}}\) with the expected resulting state \(q_1\), and drawing the sodium hydroxide solution corresponding to the input symbol \(\sigma_{\text{draw}}\) with the expected resulting state \(q_2\). \(t_6\) for monitoring audio events and recording audio-based chemical quantity estimates is assigned to the Audio Supervisor Agent, corresponding to the input symbol \(\sigma_{\text{audio\_detect}}\), used to detect titration liquid droplet events and provide audio-based volume estimates. \(t_7\) for generating a report with high-confidence data is assigned to the Summarizer Agent, corresponding to the input symbol \(\sigma_{\text{verify}}\) with the expected resulting state \(q_{\text{accept}}\). Simultaneously, the Planner Agent instantiates the statistics-based chemical quantity logger and the high-confidence chemical quantity recorder, and distributes the task allocation table to each agent, initiating the experiment execution process.

The Vision Supervisor Agent continuously monitors visual observations and pH sensor data. After confirming that the beaker is placed on the magnetic stirrer, the pH meter probe is immersed in the solution, and the rubber-headed pipette is inside the beaker, it triggers the Action Agent to execute \(t_5\). The robotic arm moves to the beaker position to grasp the rubber-headed pipette, corresponding to the input symbol \(\sigma_{\text{grasp}}\), and the state machine transitions from the initial state \(q_0\) to the state of having grasped the rubber-headed pipette \(q_1\) via the grasp completion event. The Action Agent continues to execute the drawing operation within \(t_5\), moving the pipette to the sodium hydroxide solution container to draw a sufficient amount of solution and then returning to the titration position, corresponding to the input symbol \(\sigma_{\text{draw}}\), and the state machine transitions from \(q_1\) to the state of having drawn the solution \(q_2\) via the draw completion event. When the Vision Supervisor Agent detects that the state machine has evolved to the ready-for-titration state \(q_3\), it executes subtask \(t_4\) to activate the statistics-based chemical quantity logger to perform the titration operation \(t_5\). 

The logger initiates its core three-state finite state machine: in the motion state, it controls the gripper to close at a constant rate, allowing the solution to drip drop by drop, while continuously monitoring pH changes; when a pH change is detected, it enters the waiting state, pausing action until the pH stabilizes, corresponding to the input symbol \(\sigma_{\text{stable}}\); if anomalies such as droplet failure or sensor timeout are detected, it enters the reset state to execute error correction procedures. 

In the titration experiment, the Planner Agent, during initialization, sets the gripper to close at a preset rate \(r_{\text{drop}}\) based on task characteristics, robot state, and meta-actions, and uses the statistics-based chemical quantity logger. As shown in ~\ref{contro} (b), this causes the gripper to close at a uniform rate, with the pipette steadily dispensing liquid drop by drop. The pH value is continuously monitored, and the volume of liquid dispensed is statistically recorded based on the Gripper Displacement-Volume Mapping Model:

\begin{equation}
\Delta v = f(\Delta d) = \frac{0.046875}{0.000415625} \cdot \Delta d \approx 112.78 \cdot \Delta d
\end{equation}
where \(\Delta d\) represents the displacement increment of the gripper, indicating the distance the gripper closes during the current control cycle; \(\Delta v\) represents the volume of liquid dispensed per drop (mL), corresponding to the volume of liquid dispensed when the gripper closes by \(\Delta d\); \(f\) is the gripper displacement-volume mapping function, established through offline calibration; \(0.000415625\) is the calibration constant representing the benchmark displacement corresponding to a 0.000415625 mm gripper movement; \(0.046875\) is the calibration constant (mL) representing the volume of liquid dispensed when the gripper closes 0.000415625 mm; and \(112.78\) is the conversion factor (mL/mm), calculated as \(\frac{0.046875}{0.000415625}\), indicating the volume of liquid dispensed per unit displacement.

The Planner Agent initializes the mapping relationship between gripper displacement and liquid volume at the start, and the statistics-based chemical quantity logger accumulates the total estimated volume:

\begin{equation}
\hat{V}_{\text{stat}} = \sum \Delta v = \sum f(\Delta d)
\end{equation}
where \(\hat{V}_{\text{stat}}\) represents the cumulative estimated volume of liquid dispensed from the start of the experiment to the current time (mL).

When the pH change rate exceeds a threshold or approaches the target value \(|s_c - s_c^{\text{target}}| \leq \varepsilon_{\text{pH}}\), the controller enters a waiting state, pauses the gripper movement, and holds the current position, awaiting subsequent instructions.

If the Vision Supervisor Agent and Audio Supervisor Agent detect anomalous events, such as abnormal droplet fall, sensor timeout, or operation timeout, the system executes correction processes such as retracting the gripper or re-positioning. After the correction, the system returns to the motion state and reports to the supervisor agents. The statistics-based chemical quantity logger is initialized to the **motion state** when activated, and it cycles through three states during the operation process based on chemical feedback and anomaly detection, until the Vision Supervisor Agent determines that the state machine transitions to the next state or the Vision Supervisor Agent actively terminates the process.

The Audio Supervisor Agent executes subtask \(t_6\), continuously monitoring contact-based audio signals. Upon detecting a droplet sound event, it estimates the single-drop volume (approximately 0.046875 mL) and accumulates the total volume, while providing a confidence score for each estimate, corresponding to the input symbol \(\sigma_{\text{audio\_detect}}\). The statistics-based chemical quantity logger simultaneously records statistical volume estimates based on the gripper displacement-volume mapping model. The Vision Supervisor Agent executes subtask \(t_2\), recording the pH value and corresponding volume at each stable moment, and notifies the system to enter fine-grained control mode when the pH approaches the target value of 12.5, corresponding to the input symbol \(\sigma_{\text{record}}\). The high-confidence chemical quantity recorder fuses in real time the audio-based volume estimates and their confidence from the Audio Supervisor Agent, the statistical volume estimates and their confidence from the statistics-based chemical quantity logger, and the pH values, temperature, and state transition events from the Vision Supervisor Agent, generating the most reliable volume records through confidence-weighted fusion. When the Vision Supervisor Agent detects the endpoint, it generates \(\sigma_{\text{endpoint}}\), and the state machine transitions to the titration completion state \(q_6\). After verification, it generates \(\sigma_{\text{verify}}\), and the state machine finally transitions to the titration experiment accepting state \(q_{\text{accept}}\).

When the Vision Supervisor Agent detects that the pH value has reached \(12.5 \pm 0.05\) and remained stable for more than 5 seconds, it generates the endpoint detection event \(\sigma_{\text{endpoint}}\), causing the state machine to transition from the titration in progress state \(q_4\) to the titration completion state \(q_6\). After endpoint verification, it generates the verification completion event \(\sigma_{\text{verify}}\), and the state machine finally transitions to the titration experiment accepting state \(q_{\text{accept}}\). The Summarizer Agent, triggered to execute subtask \(t_7\), generates a comprehensive experimental report based on the high-confidence data \(\mathcal{D}\). The report includes experimental metadata such as experiment time, initial conditions, and target conditions; a record of the operational procedure; a detailed data table of each drop; the equivalence point volume and half-equivalence point pH value calculated from the titration curve; and the precisely located equivalence point position verified through first and second derivative analysis. The report automatically generates four visual charts: the titration curve showing the relationship between pH and volume, the first derivative plot with a peak precisely corresponding to the equivalence point, the second derivative plot further verifying the equivalence point accuracy through zero-crossing, and an enlarged view of the transition jump region demonstrating the uniform distribution of data points near the equivalence point. If any equipment shortages, droplet anomalies, or sensor failures occur during the experiment, the system provides real-time feedback to the user regarding the specific issue and awaits further instructions.

\subsection{Acid-base titration experiment}

The embodied intelligent robot-assisted acid--base titration experiment of maleic acid is of great significance for studying the stepwise dissociation mechanism of weak organic acids and promoting the automation of precision analytical chemistry. As a typical dicarboxylic acid, the $\mathrm{p}K_{a1}$ and $\mathrm{p}K_{a2}$ values of maleic acid are key parameters in fields such as food additive detection and pharmaceutical synthesis control. Traditional manual titration faces limitations including large operational errors and low data acquisition frequency, making it difficult to accurately capture the subtle pH changes during the two-step dissociation process. In contrast, the developed embodied intelligent robot integrates common laboratory pipettes with real-time pH sensing (response time $< 0.1\,\mathrm{s}$) to achieve fully automated operation throughout the entire process. In this experiment, $25\,\mathrm{mL}$ of $0.1\,\mathrm{M}$ maleic acid (initial $\mathrm{pH} = 1.51$) was titrated with $0.1\,\mathrm{M}$ NaOH, yielding 2322 data points (volume range: $0.00$--$95.28\,\mathrm{mL}$). After denoising via Savitzky--Golay filtering, the first derivative $\frac{\mathrm{d}\,\mathrm{pH}}{\mathrm{d}V}$ and second derivative $\frac{\mathrm{d}^2\,\mathrm{pH}}{\mathrm{d}V^2}$ were calculated using numerical differentiation to locate the inflection points. The complete titration curve exhibits typical characteristics of dicarboxylic acid dissociation: in the initial stage ($0$--$20\,\mathrm{mL}$), the pH slowly rises to $2.48$; in the first inflection stage ($20$--$35\,\mathrm{mL}$), the pH sharply increases to $6.01$ (inflection point: $25.18\,\mathrm{mL}$, $\mathrm{pH} = 3.81$); in the buffer stage ($35$--$46\,\mathrm{mL}$), the pH gently rises to $6.85$; in the second inflection stage ($46$--$55\,\mathrm{mL}$), the pH rapidly jumps to $11.71$ (inflection point: $50.96\,\mathrm{mL}$, $\mathrm{pH} = 9.37$); and in the excess stage ($>55\,\mathrm{mL}$), the pH stabilizes between $11.71$ and $12.5$. The enlarged views of the inflection points show uniformly distributed data without abnormal fluctuations, highlighting the high-precision control advantage of the robot.

Derivative analysis further validates the dissociation endpoints and enables accurate $\mathrm{p}K_a$ calculation. Two distinct peaks ($1.08\,\mathrm{pH}/\mathrm{mL}$ and $2.45\,\mathrm{pH}/\mathrm{mL}$) in the first derivative plot correspond to the two inflection points, with peak intensities positively correlated with inflection amplitudes and clear separation, demonstrating high-resolution data acquisition. The second derivative plot precisely locates the inflection points through ``zero-crossing'' without interference from spurious zeros, achieving a positioning accuracy of $\pm\,0.01\,\mathrm{mL}$. Based on the principle that the pH at the half-equivalence point equals $\mathrm{p}K_a$, the calculated values are $\mathrm{p}K_{a1} = 1.91$ (corresponding to $12.58\,\mathrm{mL}$) and $\mathrm{p}K_{a2} = 6.23$ (corresponding to $38.06\,\mathrm{mL}$), with a relative error of only $0.52\%$ compared to the literature values ($\mathrm{p}K_{a1} = 1.92$, $\mathrm{p}K_{a2} = 6.23$). To account for the randomness of single experiments, three parallel experiments were conducted, yielding $\mathrm{p}K_{a1} = 1.97$ and $\mathrm{p}K_{a2} = 6.13$ with relative errors of $2.60\%$ and $1.61\%$, respectively, both significantly lower than the $5\%$ error threshold of manual titration. The 2322 high-resolution data points ensure data reliability, and the $\mathrm{p}K_a$ calculation accuracy meets the requirements of academic research. The nearly 8-hour ($7\,\mathrm{h}\,49\,\mathrm{min}\,22\,\mathrm{s}$) continuous error-free operation not only demonstrates the reliable performance of the embodied intelligent robot in prolonged experimental scenarios but also ensures the completeness and continuity of data acquisition.

To further verify the versatility of the embodied intelligent robot in different types of acid--base titrations, parallel studies on the titration of a strong acid ($\mathrm{HCl}$) and a weak acid ($\mathrm{CH_3COOH}$) with $0.1\,\mathrm{M}$ sodium hydroxide were conducted simultaneously. Complete titration curves of both acids were obtained using the same automated system (total of 2700 data points, volume range: $0.00$--$50.00\,\mathrm{mL}$), with distinct characteristics reflecting their intrinsic dissociation differences. As a strong acid that dissociates completely, $\mathrm{HCl}$ exhibits no buffer platform in its titration curve, with the inflection point located at $25.29\,\mathrm{mL}$ ($\mathrm{pH} = 6.98$), representing a deviation of only $1.16\%$ from the theoretical equivalence point ($25.00\,\mathrm{mL}$). In contrast, $\mathrm{CH_3COOH}$, as a weak acid undergoing partial dissociation ($\mathrm{CH_3COOH \rightleftharpoons CH_3COO^- + H^+}$), shows different features: the initial pH slowly rises from $2.9$ to $4.5$, followed by a buffer platform ($4.5$--$6.0$) in the $9$--$25\,\mathrm{mL}$ range due to the $\mathrm{CH_3COOH/CH_3COO^-}$ buffer pair (governed by the Henderson--Hasselbalch equation); in the $25$--$26\,\mathrm{mL}$ range, the pH increases from $6.0$ to $11.0$, with the inflection point located at $25.52\,\mathrm{mL}$ ($\mathrm{pH} = 8.18$, where the equivalence point pH is alkaline due to the hydrolysis of sodium acetate). In the excess stage, both acids exhibit a stable pH between $11.5$ and $12.5$. These results complement the technical validation of maleic acid titration, providing more comprehensive automated technical support for the qualitative identification and quantitative analysis of acidic substances.

To quantitatively verify the repeatability and data validity of the embodied intelligent robot titration system, the standard deviation between the measured pH values and theoretical simulation values at each volume point was calculated based on three parallel titration experiments of $0.1\,\mathrm{M}$ $\mathrm{HCl}$ with $0.1\,\mathrm{M}$ $\mathrm{NaOH}$. The results show an overall standard deviation of $0.01$--$0.10$ (excluding the inflection range) with distinct interval characteristics: in the initial stage ($0$--$23\,\mathrm{mL}$, excess strong acid), the standard deviation stabilizes at $0.01$--$0.10$; in the inflection range ($23$--$27\,\mathrm{mL}$), the standard deviation increases to $5.47$ (attributed to the extreme sensitivity of pH to titrant volume in this range, leading to high uncertainty); and in the excess stage ($>27\,\mathrm{mL}$, excess strong base), the standard deviation falls back to $0.01$--$0.10$, showing excellent agreement with theoretical values. Meanwhile, the minimal inter-parallel experimental error reflects the stable consistency of the robot in liquid delivery, pH sensing, and data recording, confirming the system's repeatability and reliability from a statistical perspective.

\begin{figure}[h]
\centering
\includegraphics[width=\textwidth]{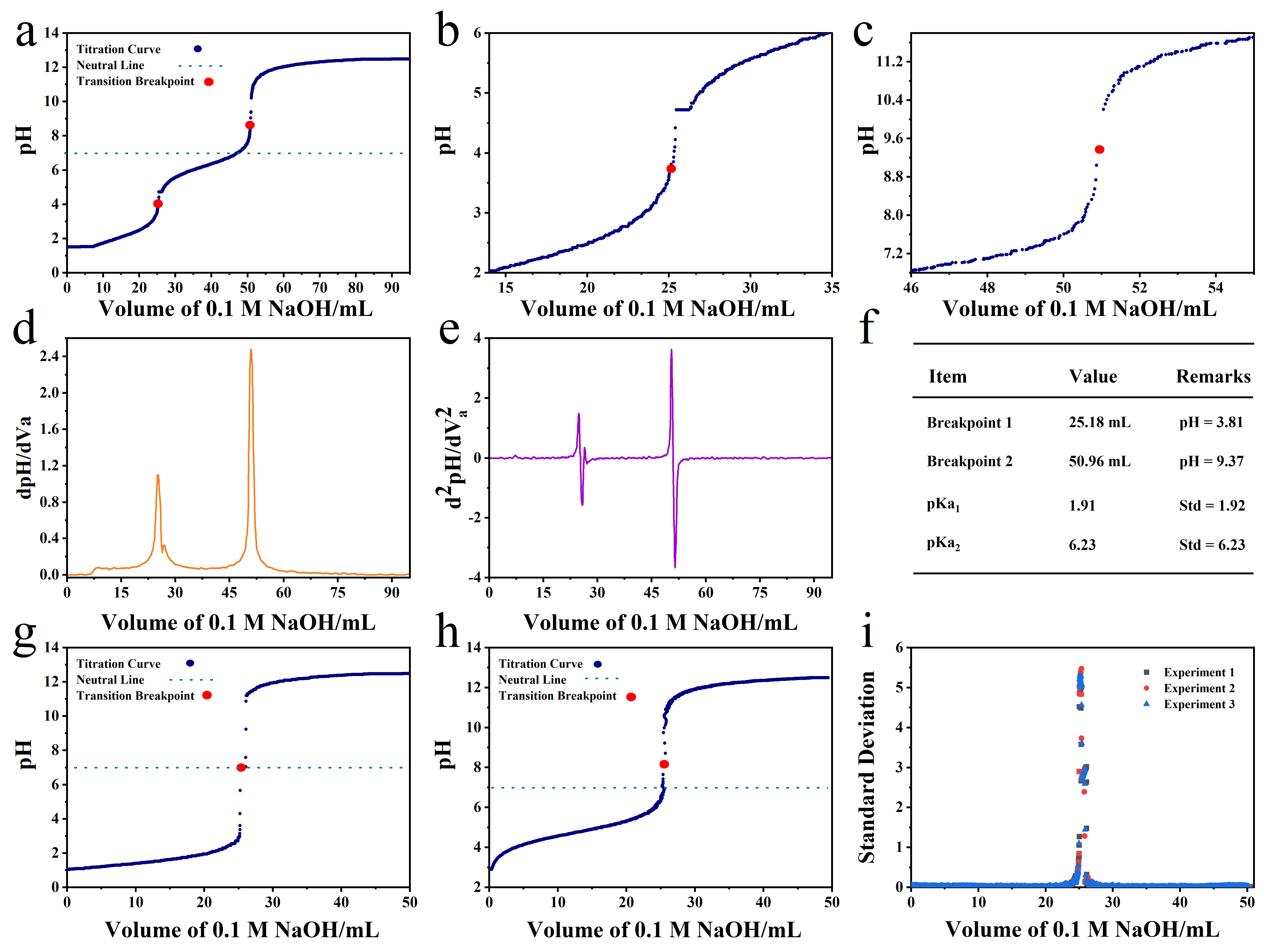}
\caption{\textbf{Acid-Base Titration Experiments.} \textbf{a} Titration curve of maleic acid (concentration: $0.1\,\mathrm{M}$, initial volume: $25\,\mathrm{mL}$, initial $\mathrm{pH}=1.51$). \textbf{b-c} Enlarged view of the transition region in the maleic acid titration curve. \textbf{d-e} First-derivative and second-derivative plots of the maleic acid titration curve. \textbf{f} Key conclusions from the Agent data analysis report for the maleic acid titration experiment. \textbf{g-h} Titration curves of hydrochloric acid and acetic acid (both at $0.1\,\mathrm{M}$ concentration, initial volume: $25\,\mathrm{mL}$). \textbf{i} Standard deviation plot comparing the hydrochloric acid titration curve with theoretical values (comparison of three parallel experiments and theoretical simulation).
}
\label{fig2}
\end{figure}

\subsection{Complexometric titration experiment}
To verify the reliability of the color perception channel of the embodied intelligent robot in complexometric titration, the EDTA complexometric titration method was employed for the determination of calcium ion ($\mathrm{Ca^{2+}}$) concentration, with calconcarboxylic acid (calcium indicator) used as the color-change reporter. During the experiment, the robot synchronously collected color channel data and titrant volume data, while accurately tracking the color characteristics of the reaction system: in the initial stage, the solution turned magenta after the addition of the calcium indicator (confirming the formation of the $\mathrm{Ca^{2+}}$--indicator complex); as the $0.01\,\mathrm{M}$ EDTA standard solution was continuously added, the robot captured subtle color variations. When the titrant volume reached $2.5\,\mathrm{mL}$, the color perception channel clearly identified the abrupt color transition of the solution from fuchsin red to sapphire---an observation that indicates the complete chelation of $\mathrm{Ca^{2+}}$ by EDTA to form the stable $\mathrm{Ca}$--EDTA complex. Thereafter, the solution color remained stable without any false color signals detected. Quantitative analysis based on 3 parallel experiments showed that the $\mathrm{Ca^{2+}}$ concentration determined by the robot was $0.00968\,\mathrm{M}$, with a relative error of only $3.20\%$ compared to the standard concentration of $0.0100\,\mathrm{M}$, which meets the experimental precision requirements. These results fully demonstrate that the robot can not only accurately identify the critical color-transition endpoint in the EDTA--$\mathrm{Ca^{2+}}$ complexometric titration but also exhibit high stability and reliability in its color perception chemical channel. This effectively avoids the subjective errors associated with manual endpoint judgment and provides technical support for the automation of analytical chemistry experiments that rely on color recognition.

\begin{figure}[!t]
\centering
\includegraphics[width=\textwidth]{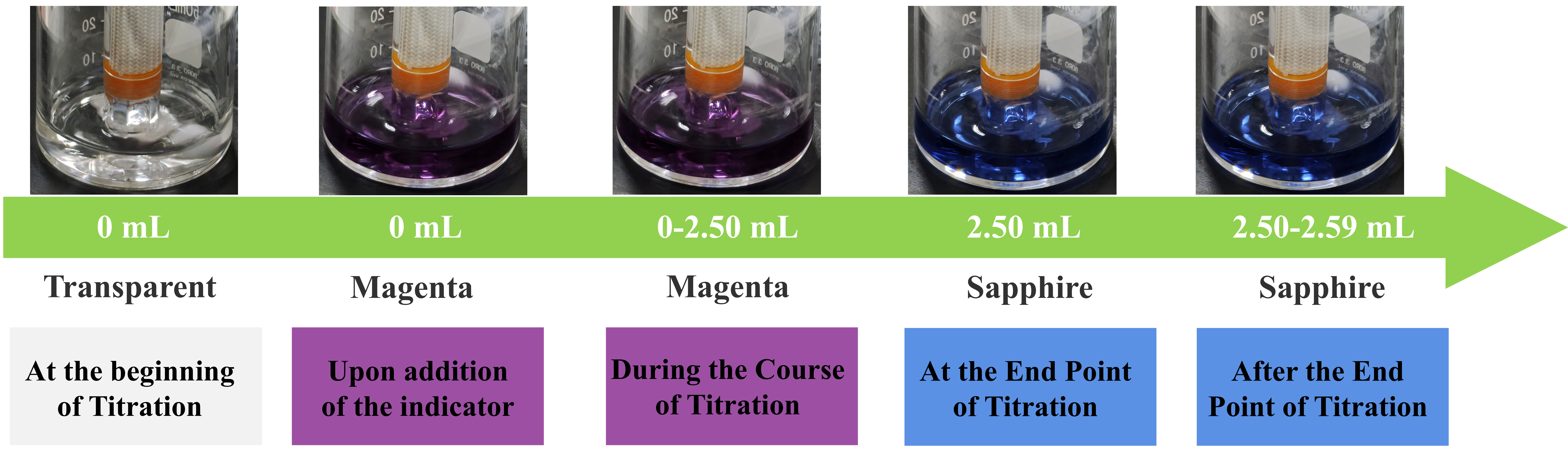}
\caption{\textbf{Chemosensor with Color Perception.} The color transition sequence (from left to right) during the complexometric titration of $\mathrm{Ca^{2+}}$ concentration using a $0.02\,\mathrm{M}$ EDTA standard solution against a $0.01\,\mathrm{M}$ $\mathrm{CaCl_2}$ solution (initial volume $5\,\mathrm{mL}$) with a calcium indicator was as follows: the solution initially exhibits a magenta color due to the calcium--indicator complex, and finally turns sapphire at the endpoint upon the release of the free indicator.
}
\label{fig3}
\end{figure}

\subsection{Solid Weighing and Dissolution Experiment}

To validate the generalization ability and task-level adaptability of AgentChemist in diversified chemical experimental tasks, we conducted experiments on solid weighing and dissolution tasks in addition to acid-base titration.

Task description: Weigh a specified mass of NaCl solid (e.g., 2.50 g), transfer it to a beaker, add a fixed amount of solvent (50 mL deionized water), start the magnetic stirrer until the solid is completely dissolved, and record the weighing error and dissolution time. Evaluation metrics include weighing absolute error, dissolution completion time, and operation success rate. The expected results are a weighing error of \(\leq 0.12\) g, autonomous completion of the dissolution process, and a success rate of over 95\%.

\begin{table}[h]
\centering
\caption{Experimental Results for Solid Weighing and Dissolution Task}
\label{tab:solid_task}
\small
\begin{tabular}{c|c|c|c}
\toprule
\textbf{No.} & \textbf{Weighing Error (g)} & \textbf{Time (s)} & \textbf{Success} \\
\midrule
1 & 0.08 & 45.2 & Yes \\
2 & 0.11 & 47.8 & Yes \\
3 & 0.05 & 43.5 & Yes \\
4 & 0.15 & 52.1 & Yes \\
5 & 0.09 & 46.3 & Yes \\
6 & 0.07 & 44.9 & Yes \\
7 & 0.12 & 48.6 & Yes \\
8 & 0.10 & 47.2 & Yes \\
9 & 0.13 & 49.5 & Yes \\
10 & 0.06 & 45.8 & Yes \\
\midrule
\textbf{Average} & 0.096 & 47.09 & 100\% \\
\textbf{Std Dev} & 0.032 & 2.61 & - \\
\textbf{Target} & \(\leq 0.12\) & - & \(\geq 95\%\) \\
\bottomrule
\end{tabular}
\end{table}

\subsection{Generalization Verification of AgentChemist}

We evaluate the generalization capability of the AgentChemist framework from two dimensions. First, we validate the system's adaptability to different types of chemical experiments and experimental environments through a solid drug weighing task. Second, we verify the framework's robustness and compatibility with changes in the underlying model by replacing the large language model of the agents.

\subsubsection{Task and Environment Generalization: Solution Preparation Experiment}
To verify the generalization capability of AgentChemist in different types of chemical experiments, we designed a solution preparation task, which differs significantly from the previously mentioned titration experiment in terms of action patterns, chemical perception requirements, and execution accuracy. The experiment requires the robot to weigh a specified mass of NaCl solid from a reagent bottle, transfer it to a medium-sized beaker containing 50 mL of deionized water, and activate the magnetic stirrer. We use weighing error as the evaluation metric.
To verify the environment generalization, in addition to the standard layout where instruments are arranged neatly in preset positions, we also conducted the experiment under a disturbed layout, with small, medium, and large beakers placed in the setup.

We conducted 50 tests, and the experimental results are shown in Table~\ref{tab:generalization_solution}. Under the standard layout, AgentChemist achieved a task success rate of 98\% with an average weighing error of 0.08 g, demonstrating reliable execution in a conventional experimental environment. When the environmental complexity increased to a disturbed layout containing small, medium, and large beakers, the system still maintained a success rate of 94\%, with the average weighing error only slightly increasing to 0.11 g. These results indicate that although the presence of multiple beakers of different sizes introduces ambiguity in visual perception, AgentChemist is still able to effectively distinguish the target beaker from non-target objects and maintain a high level of operational precision. This experiment validates the generalization robustness of the AgentChemist framework when facing different task types and common variations in laboratory environments.

\begin{table}[h]
\centering
\caption{Generalization Performance of the Solution Preparation Task in Different Environments}
\label{tab:generalization_solution}
\begin{tabular}{ccc}
\toprule
\textbf{Environmental Conditions} & \textbf{Success Rate} & \textbf{Average Weighing Error (g)} \\
\midrule
Standard Layout & 98\% & 0.08 \\
Disturbed Layout (Multiple Sizes of Beakers) & 94\% & 0.11 \\
\bottomrule
\end{tabular}
\end{table}

\subsubsection{Framework Generalization: LLM Replacement Experiment}

To verify the compatibility and generalization ability of the AgentChemist framework with different underlying models, we replaced the base models of the agents while keeping the framework structure unchanged, and conducted a gradient-based comparative experiment ranging from high-performance baselines to cross-technical paradigms. In the experimental setup, the control group uses the default Qwen3 series model as the performance benchmark. In Experiment Group A, each agent is replaced with an earlier Qwen2 series model to test the framework’s downward compatibility with older model versions. Experiment Group B completely departs from the large language model (LLM) paradigm and replaces the agents with a traditional non-LLM model combination, including a rule-based task parser, a ResNet50 object detection network, an Audio-CLIP audio encoder, and a template-filling script. The results are shown in Table~\ref{tab:llm_generalization}.

\begin{table}[h]
\centering
\caption{System Generalization Performance under Different Model Configurations}
\label{tab:llm_generalization}
\small
\begin{tabular}{cccc}
\toprule
\textbf{Configuration} & \textbf{Task Completion Rate} & \textbf{pH Deviation} & \textbf{Anomaly Detection Accuracy} \\
\midrule
Control Group & 96\% & 0.11 & 94\% \\
Experiment Group A & 92\% & 0.15 & 89\% \\
Experiment Group B & 78\% & 0.26 & 73\% \\
\bottomrule
\end{tabular}
\end{table}

The setup of Experiment Group A aims to test the framework’s downward compatibility with earlier versions of the same series models. Compared to the Qwen3 series, the Qwen2 series has certain gaps in semantic understanding, multimodal alignment, and reasoning efficiency. A moderate performance decrease was expected, which would verify if the framework can accommodate model versions with different capabilities. This would allow users to flexibly choose a compatible model version based on actual resource limitations without modifying the framework code. The setup of Experiment Group B, on the other hand, aims to perform a cross-technical paradigm stress test. In this group, the Planner Agent uses a rule-based task parser to parse user instructions through predefined syntax rules and keyword matching. The Vision Supervisor Agent uses a ResNet50 object detection network to identify the position and status of experimental instruments and reagents. The Audio Supervisor Agent uses Audio-CLIP to map contact audio to a joint embedding space aligned with vision and text, enabling the detection and classification of acoustic events. The Summarizer Agent uses a template-filling script to populate experimental data into a pre-set report template. This combination completely departs from the LLM paradigm and aims to verify whether the core collaborative mechanism of the framework can still be compatible with traditional AI methods.

The experimental results show that after replacing with the earlier Qwen2 series models, Experiment Group A experienced a reasonable performance drop within an expected range: task completion rate dropped from 96\% to 92\%, pH deviation increased from 0.11 to 0.15, anomaly detection accuracy decreased from 94\% to 89\%, and time increased from 238 seconds to 251 seconds. Despite the performance drop, the system maintained a high task completion rate and operational precision, validating the framework’s good downward compatibility. Users can choose to deploy older model versions in resource-limited scenarios, and the framework’s collaborative mechanism still works reliably. Although the performance of Experiment Group B significantly dropped, with a task completion rate of 78\%, pH deviation of 0.26, anomaly detection accuracy of 73\%, and time increasing to 298 seconds, the system was still able to maintain basic operations and successfully complete most experimental steps. Considering that this group completely departs from the LLM technology paradigm and uses traditional methods such as rule-based parsing, convolutional neural networks, and contrastive learning encoders, this result strongly demonstrates the architecture neutrality of the AgentChemist framework. Its core value lies not only in adapting to large language models but also in providing a universal multi-agent collaborative mechanism that is inclusive of various technical paradigms and model capabilities. Even in a non-LLM era, such an architectural design can still operate effectively, laying a solid foundation for the long-term development and cross-era compatibility of the framework. 
\section{Discussion}

The application of embodied intelligent robots in chemical experiments marks a paradigm shift in laboratory automation, from "scripted automation" to "embodied perception and autonomous decision-making intelligence." Traditional automation systems are limited by rigid processes, making it difficult to adapt to the inherent diversity and uncertainty of chemical experiments. AgentChemist integrates the semantic understanding capabilities of large language models with the physical embodiment execution capabilities of robots through a multi-agent collaborative architecture. This integration achieves a fully automated closed-loop process from natural language instructions to structured experimental reports. This capability not only significantly improves experimental efficiency but also provides a reproducible, traceable, and interpretable digital experimental paradigm, paving the way for the knowledge accumulation and sharing of scientific discoveries.

Chemical experimental scenarios present unique and challenging requirements for embodied intelligent systems. The first challenge is the ambiguity of perception. Factors such as transparent containers, colorless liquids, and lighting changes often cause pure visual perception to fail in chemical scenarios. AgentChemist compensates for the limitations of single-modality perception by introducing contact audio as a physically invariant clue to event existence, employing a multimodal fusion strategy. However, the robustness of perception in transparent and colorless scenarios still has room for improvement. The second challenge is the precision of operations. Chemical experiments require ultra-precise liquid handling 
at the milliliter level, far exceeding the precision needed for everyday robotic operations. Human remote-control demonstration data cannot cover such precise tasks, and traditional control methods are inadequate for the diversity of tasks. AgentChemist's hierarchical control architecture offers an effective solution, though factors like gripper wear and variations in liquid viscosity still impact the precision stability over long-term operation. Furthermore, real chemical laboratories exhibit highly unstructured characteristics, with challenges such as reagent position changes, equipment model differences, and varying lighting conditions. The system's generalization ability remains challenged, and the current system relies on pre-set instrument layouts and calibration parameters, meaning it is still distant from "plug-and-play" deployment. Lastly, fine-tuning large language models in vertical fields requires high-quality specialized data, and the cost of collecting chemical experimental data and annotating it is high. How to achieve stronger cross-task generalization ability with fewer demonstration data is a core issue that embodied intelligence in chemistry must address.

These challenges also present significant technical opportunities. Deepening multimodal perception will enable robots to achieve chemical scene understanding beyond human capabilities. In addition to visual and audio modalities, the introduction of tactile, thermal, and electrochemical sensors will create "super-visual" capabilities for chemical scenarios. Exploration of precision operation meta-learning will package precision control strategies into reusable "operation primitives," allowing VLA models to learn when to invoke these primitives instead of how to execute them, potentially greatly reducing data requirements for precision operations. The transfer from simulation to reality, combined with high-precision chemical simulation environments, will allow for large-scale strategy pre-training in virtual spaces, followed by fine-tuning with a small amount of real data, which may be the key path to overcoming the data bottleneck. The new paradigm of human-robot collaboration will position the robot as a "research partner" rather than just an "execution tool," requesting human guidance in uncertain steps and autonomously executing in standardized steps, thus forming a human-robot collaborative intelligent loop.

Although AgentChemist has demonstrated its effectiveness in multiple chemical experimental tasks, the current system still has several limitations. In terms of perception, the reliability of visual perception significantly decreases in scenarios with transparent containers and colorless liquids. Audio perception is prone to false positives in environments with high noise levels, and a self-calibration mechanism for long-term chemical sensor drift has not yet been established. On the operation side, the gripper displacement-to-volume mapping model of the \textcolor{yellow}{precision controller} relies on offline calibration and cannot adapt to variations in liquid viscosity or differences in pipette models. The success rate of solid handling is lower than that of liquid handling, reflecting insufficient modeling of the physical properties of particulate matter. On the deployment side, the system is sensitive to environmental factors such as experimental table layout, instrument models, and lighting conditions. Hardware costs are high, and the software environment relies on specific versions of ROS and deep learning frameworks, making environmental migration costly. Regarding data, the pretraining of VLA models consumes substantial computational resources, and vertical domain fine-tuning still requires hundreds of demonstration data. Anomaly detection performance is insufficient in scenarios requiring domain-specific prior knowledge. In terms of interaction, the accuracy of natural language instruction parsing is limited by the ambiguity and vagueness of the instructions, and the system lacks an active clarification mechanism for user intent.

Looking towards the long-term development of chemical experiment automation, future breakthroughs should be made in several directions. In terms of perception enhancement, more physical and chemical sensing channels should be integrated, and active perception strategies should be developed, allowing robots to autonomously choose "where to look, where to listen, and where to measure" to eliminate uncertainties in key states. In terms of operational expansion, common chemical operations should be packaged into parameterized operation primitives, with each primitive incorporating closed-loop control strategies and anomaly handling logic, allowing the VLA model to learn the timing of primitive invocation rather than the low-level action generation. For ease of deployment, self-calibration and adaptive technologies should be developed, combined with digital twin simulations of experimental processes, promoting containerized deployment and standardization of hardware abstraction layers. In terms of efficiency improvement, the system should evolve from "autonomous execution" to "autonomous planning," developing closed-loop task re-planning capabilities, and exploring multi-robot collaborative models to enhance experimental throughput. For human-robot interaction, a multi-turn dialog-based instruction clarification mechanism should be developed, introducing demonstration learning and feedback learning, so that non-expert users can teach the robot new tasks through a small number of demonstrations, ultimately achieving a "human-robot research partner" relationship. In terms of data dependence, domain-specific prior knowledge should be embedded in model training, and transfer learning from simulation to reality should be developed, generating vast synthetic data in chemical simulation environments to overcome the data collection bottleneck.

The exploration of AgentChemist demonstrates that embodied intelligence is opening new possibilities for chemical experiment automation. From automation to intelligence, from single-task to multi-scene generalization, and from rigid execution to adaptive decision-making, we are standing at the starting point of a paradigm shift in laboratories. With continuous breakthroughs in perception enhancement, operational generalization, deployment simplification, efficiency improvement, interaction optimization, and data efficiency, embodied intelligent robots are expected to truly become intelligent partners for chemical researchers, jointly exploring the unknown territories of the material world.

\section{Methods}\label{method}




\subsection{Platform Hardware Configuration}
We have developed AgentChemist on the AgileX Mobile
ALOHA  robot (COBOT Magic, AgileX Robotics Co. Ltd., Dongguan, China). 
AgileX Mobile
ALOHA  robot combines a wheeled mobile base with two 7-degree-of-freedom (DoF) robotic arms to enable flexible navigation and precise manipulation in real laboratory environments.

\begin{figure}[!t]
\centering
\includegraphics[width=\textwidth]{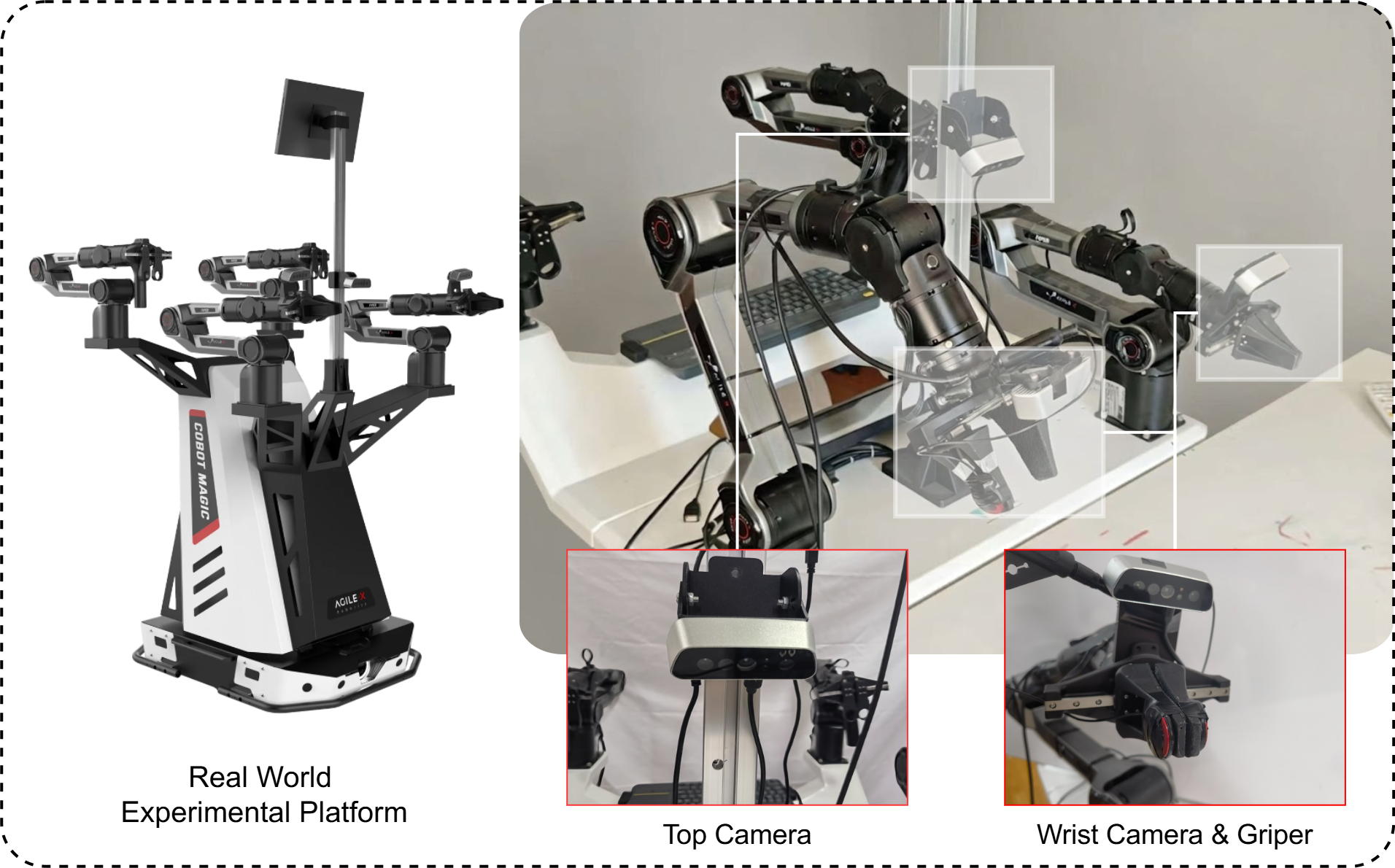}
\caption{Configuration of the AgentChemist robotic experimental platform. The system features a Tracer 2.0 mobile base with all-terrain design and independent suspension for precise navigation. Two 7-DoF robotic arms (Piper Standard) are mounted on a rigid frame, offering 1.5 kg payload capacity and ±0.1 mm repeatability for precise laboratory manipulation. The perception system includes three Orbbec Dabai DC1 depth cameras (1920×1080, 30 fps) mounted on the wrists and front, plus two contact microphones embedded in the grippers to capture acoustic signatures during experimental operations.
}
\label{platform}
\end{figure}

Fig~\ref{platform} illustrates the configuration of the robotic system.
The mobile base is a Tracer 2.0 two‑wheel differential drive chassis (Tracer 2.0, AgileX Robotics Co. Ltd., Dongguan, China), featuring a rugged all‑terrain design with 4×4 independent suspension. The differential drive configuration provides omnidirectional movement capability through coordinated wheel speed control, enabling precise positioning in confined laboratory spaces and smooth navigation around benchtops and equipment rack. 
Two 7-DoF robotic arms (Piper Standard model) (Piper, AgileX Robotics Co. Ltd., Dongguan, China) are mounted on a rigid aluminum frame ($1125\times758\times1507mm$) connected to a mobile base. Each arm has a working range of 626mm and a payload capacity of 1.5kg, which exceeds the requirements for handling common laboratory glassware, reagent bottles, and instruments. The maximum gripping range of the gripper is 80mm, and the maximum gripping force is 10NM. The arms have a repeatability of ±0.1mm, making them adaptable to the vast majority of chemical laboratory settings.
The perception system of AgentChemist is designed to provide comprehensive environmental awareness and detailed monitoring of experimental processes through a multi-modal sensor configuration.
AgileX Mobile
ALOHA  robot is equipped with three Orbbec Dabai DC1 depth cameras (Orbbec Dabai DC1, Orbbec Inc., Shenzhen, China) on the left wrist, right wrist, and front, respectively. These cameras capture RGB images in real time at a resolution of 1920×1080 and a frame rate of 30 fps, with an average transmission latency of 30–45 milliseconds, providing stable and reliable image input for AgentChemist's visual chemical perception tasks. 2 contact microphones (AOM‑5024L‑HDR‑R, PUI Audio Inc., Fairborn, America) are embedded within  grippers to capture vibration and acoustic signatures during liquid dispensing, solid pouring, and mechanical contact events.

\subsection{Software Implementation}

The robot's low-level control communicates with the chassis and robotic arm via the CAN bus. The system uses the SocketCAN framework to send and receive CAN frames through the can0 interface, with a data transmission rate of 1 Mbps. The robotic arm control commands (gripper displacement, joint positions) and state feedback (current position, speed) are all encapsulated as ROS messages. The software environment of the AgentChemist  is built on the Ubuntu 20.04 LTS operating system, with ROS~\cite{quigley2009ros}  (Robot Operating System) serving as the core communication middleware. The system is primarily implemented in Python 3.10.19, and the development environment uses Visual Studio Code (version 1.208.2). It integrates modules for large language model inference, robot control, multimodal perception, and multi-agent collaboration. The system adopts a distributed and modular architecture, supporting the localized deployment and efficient inference of large language models such as Qwen3-VL~\cite{qwen3-vl}and Qwen2-Audio~\cite{qwen2-audio}.

\subsection{Training Methods and Prompt Design for Agents}

The agents in the AgentChemist framework are built upon different pre-trained large language models and driven by structured prompts to guide their behavior. This section provides a detailed explanation of the training methods for each agent, the design of the prompt templates, and the model fine-tuning strategies.

\subsubsection{Planner Agent Prompt Template}

The operation of the Planner Agent is driven by a structured prompt that encodes the system architecture, task requirements, and output format, enabling the large language model to transform natural language instructions into structured executable workflows. The complete prompt template is as follows:

\begin{quote}
\small
``You are the Planner Agent within AgentChemist, a multi-agent robotic system designed for autonomous chemical experiments. You must understand the architecture of this system. AgentChemist consists of the following agents and modules:

\begin{itemize}
    \item \textbf{Vision Supervisor Agent}: A visual observer that continuously monitors visual observations of the experimental process and chemical sensor data to determine the current state of the Finite State Machine (FSM). It tracks the evolution of the FSM, decides when to activate actuators, supervises the experimental progress, and reports abnormal states to you.
    \item \textbf{Audio Supervisor Agent}: An observer that takes contact-based audio from the environment as input. It provides chemical quantity estimates based on acoustic signals and monitors the state of the FSM according to acoustic events.
    \item \textbf{Action Agent}: An actuator responsible for driving the robot to execute operational primitives.
    \item \textbf{Statistics-based Chemical Quantity Logger}: Activated by the Vision Supervisor Agent when a predicted action is expected to cause a change in chemical quantities. It outputs actions for the robot to execute at a constant speed and provides chemical quantity estimates based on statistical analysis (e.g., gripper displacement-to-chemical-quantity mapping).
    \item \textbf{High-Confidence Chemical Quantity Recorder}: A central recording mechanism that receives statistical estimates, audio estimates, state transitions, and anomaly events. It fuses multi-modal data to generate the most reliable chemical quantity records.
    \item \textbf{Summarizer Agent}: Activated upon completion of the experiment to generate a final structured report based on high-confidence data, in accordance with user requirements.
\end{itemize}

Your inputs are: the user instruction \(\{instruction\}\), visual observations \(\{image\_observations\}\), and the initial robot state \(\{robot\_state\}\).

You must strictly perform the following operations:

\textbf{1. Semantic Parsing:} Identify key information from the instruction \(i \in \mathcal{I}\) and observations \(o_v \in \mathcal{O}\):
\begin{itemize}
    \item Experimental entities (objects, instruments, reagents)
    \item Initial conditions (chemical quantity = \(x_o\) unit)
    \item Target conditions (chemical quantity = \(x_e\) unit)
    \item Operation sequence intent
    \item Data recording requirements
    \item Expected output format
    \item Operation primitives: atomic robot actions
\end{itemize}

\textbf{2. Environment Verification:} After semantic parsing, compare the natural language instruction with the visual observations to determine whether the current environment meets the experimental requirements:
\begin{itemize}
    \item If the requirements are satisfied, proceed to state machine initialization.
    \item If the requirements are not satisfied, provide feedback to the user using the following template: \\
    \textit{``The experiment you wish to complete is [experiment name]. The required instruments and reagents are [instrument names] [reagent names]. The current experimental scene lacks [missing object names].''}
\end{itemize}

\textbf{3. State Machine Initialization:} Initialize the experiment as a finite state machine \(S\):
\[
S = (Q, \Sigma, \delta, q_0, F)
\]
where:
\begin{itemize}
    \item \(Q\): the set of possible experimental states (structured natural language)
    \item \(\Sigma\): input symbols representing agent actions and events
    \item \(\delta: Q \times \Sigma \rightarrow Q\): the state transition function, defining how the system moves from one state to another based on inputs
    \item \(q_0\): the initial state derived from the [initial conditions] in semantic parsing
    \item \(F\): the accepting states derived from the [target conditions] in semantic parsing (experiment completion conditions)
\end{itemize}

\textbf{4. Task Decomposition and Allocation:} Based on the finite state machine \(S = (Q, \Sigma, \delta, q_0, F)\), decompose the experiment into a series of atomic subtasks \(\mathcal{T} = \{t_1, t_2, \ldots, t_n\}\) and assign them to the Vision Supervisor Agent, Audio Supervisor Agent, Action Agent, and Summarizer Agent. Simultaneously, instantiate the statistics-based chemical quantity logger and the high-confidence chemical quantity recorder.

\textbf{5. Output Format:} Output a JSON object containing:
\begin{enumerate}
    \item \textbf{parsed\_instruction}: Semantic parsing results, including entities, initial conditions, target conditions, operation intent, data requirements, and output format
    \item \textbf{environment\_check}: Environment verification results, including satisfaction status, missing objects, and user feedback
    \item \textbf{state\_machine}: Complete state machine definition, including state set, input symbols, transition function, initial state, and accepting states
    \item \textbf{task\_decomposition}: Task decomposition results, including subtask list with assigned receivers, dependencies, state machine associations, and expected state transitions
    \item \textbf{module\_instantiation}: Module instantiation configuration, including activation conditions and parameters for the statistics-based chemical quantity logger and the high-confidence chemical quantity recorder
\end{enumerate}
\end{quote}

\subsubsection{Vision Supervisor Agent Prompt Template}

The operation of the Vision Supervisor Agent is driven by the following structured prompt:

\begin{quote}
\small
``You are the Vision Supervisor Agent in AgentChemist. The Planner Agent has sent you the [state\_machine] and [task\_decomposition]. You need to parse the [task\_decomposition], identify your subtasks, and use the visual observation stream and reports from other agents to assess the ongoing experimental events. Align these experimental events with the state definitions in the finite state machine [state\_machine] to determine the current experimental state \(s_c \in Q\).

\textbf{Initialization Phase:} Execute your subtasks as specified in the [task\_decomposition]. These subtasks will provide you with the initial state \(q_0\), accepting states \(F\), and the conditions under which certain actuators should be activated.

\textbf{Execution Phase:} You need to continuously supervise the experimental process. Other agents will report the completion status of their respective subtasks to you. You must confirm their execution and sensor states and bind these events to the experimental state to drive the evolution of the finite state machine:
\begin{enumerate}
    \item Infer the next state transition \(\delta(q_t, \sigma)\) and the target state \(q_{t+1} \in Q\) based on reports from other agents.
    \item Send adjustment commands \(\sigma \in \Sigma\) to the Action Agent or the statistics-based chemical quantity logger.
    \item Record chemical quantities to the high-confidence chemical quantity recorder.
    \item Maintain continuous closed-loop control until \(s_c \in F\).
\end{enumerate}

If anomalies are detected during monitoring or reported by other agents, record the anomaly information in the experimental data \(\mathcal{D}\) and send the anomaly to the Planner Agent. The Planner Agent will provide feedback to the user and await new planning instructions.

Throughout the execution of the experiment, continuously send the following to the high-confidence chemical quantity recorder:
\begin{itemize}
    \item Timestamps
    \item Chemical quantities
    \item Current state \(q_t\)
    \item State transition events
    \item Anomalous events
    \item Adjustment commands \(\sigma \in \Sigma\)
\end{itemize}

\textbf{Completion Phase:} When monitoring determines that the experiment has reached \(F \subseteq Q\), notify the Planner Agent that the task is complete.''
\end{quote}

This prompt template encodes the complete decision-making logic of the Vision Supervisor Agent, enabling it to continuously supervise the experimental process, drive state machine evolution, and manage anomalies based on the task assignment and state machine definitions from the Planner Agent. Using multimodal observations and inter-agent communication, the Vision Supervisor Agent ensures the experiment proceeds smoothly according to the expected process.

\subsubsection{Fine-tuning and Prompt Template of Audio Supervisor Agent}
\begin{quote}
\small
``You are the Audio Supervisor Agent in AgentChemist. The Planner Agent has sent you the [state\_machine] and [task\_decomposition]. You need to parse the [task\_decomposition] to identify your subtasks. You will receive contact-based audio observations \(o_a \in \mathcal{O}\), and align them with discrete subtask descriptions across modalities to determine whether the current audio segment corresponds to the type of subtask that should be executed at this step.

\textbf{Initialization Phase:} Execute the listening subtasks specified in the [task\_decomposition], identifying the types of audio events to listen for and the expected acoustic patterns at different experimental stages.

\textbf{Execution Phase:} You need to continuously process the audio stream and perform the following operations:
\begin{enumerate}
    \item \textbf{Event Detection and Segmentation:} Detect and precisely segment individual events in the continuous audio stream, estimating their start and end timestamps.
    \item \textbf{State Alignment and Validation:} Use the contextual information in the state representation \(s_c\) to logically validate the detected audio events—if the corresponding acoustic signal is detected during the execution of a subtask expected to generate that event, mark it as an "expected event"; if the acoustic signal is detected without a corresponding operational instruction, mark it as an "anomalous event."
    \item \textbf{Chemical Quantity Estimation:} Perform quantitative chemical quantity estimation based on acoustic features:
    \begin{itemize}
        \item Estimate the chemical quantity change \(\hat{q}_{\text{event}}\) corresponding to each operation
        \item Accumulate the total chemical quantity \(\hat{Q}_{\text{total}} = \sum \hat{q}_{\text{event}}\)
        \item Provide a confidence score \(\gamma \in [0,1]\) for each estimate
    \end{itemize}
    \item \textbf{State Transition Trigger:} Based on the detection results, generate input symbols \(\sigma \in \Sigma\):
    \begin{itemize}
        \item Expected event \(\rightarrow\) generate \(\sigma_{\text{event\_success}}\), driving \(\delta(q_t, \sigma_{\text{event\_success}}) = q_{t+1}\)
        \item Anomalous event \(\rightarrow\) generate \(\sigma_{\text{event\_abnormal}}\), triggering the anomaly handling state
        \item Failure to detect expected event for a long time \(\rightarrow\) generate \(\sigma_{\text{event\_timeout}}\), triggering a retry or report
    \end{itemize}
\end{enumerate}

\textbf{Anomaly Handling:} If an anomalous event is detected, record the anomaly information in the experimental data \(\mathcal{D}\) and send the anomaly to the Vision Supervisor Agent; if the anomaly exceeds the system’s autonomous handling capacity, report to the Planner Agent for re-planning.

Throughout the experiment execution, continuously send the following to the high-confidence chemical quantity recorder:
\begin{itemize}
    \item Timestamps
    \item Audio-based chemical quantity estimates \(\hat{Q}_{\text{audio}}\) and confidence scores \(\gamma\)
    \item Current state \(q_t\)
    \item Detected event types and timestamps
    \item Anomalous event records
    \item Generated state transition symbols \(\sigma \in \Sigma\)
\end{itemize}

\textbf{Completion Phase:} When the state machine reaches the accepting state \(F \subseteq Q\), confirm the completion of chemical quantity recording and stop the operation.''
\end{quote}

\subsubsection{Summarizer Agent Prompt Template}

The Summarizer Agent is activated at the completion of the experiment and is driven by the following prompt:

\begin{quote}
\small
``You are the Summarizer Agent in AgentChemist. The experiment has been completed, and the state machine has reached \(F \subseteq Q\). You have received:

\begin{itemize}
    \item State transition history: \(\{(q_0, \sigma_1, q_1), \ldots, (q_{n-1}, \sigma_n, q_n)\}\)
    \item Experimental data \(\mathcal{D}\) (timestamps, chemical quantities, anomaly flags)
    \item Subtask set \(\mathcal{T}\) from the Planner Agent
\end{itemize}

\textbf{Task:} Generate a comprehensive experimental report.

\textbf{Step 1 - Anomaly Review:} Check the state transition history and \(\mathcal{D}\) for any anomaly flags (sensor timeouts, operation failures, manual interventions).

\textbf{Step 2 - Report Generation:} Based on \(\mathcal{T}\) and \(\mathcal{D}\), generate a report containing the following, as well as [user instructions]: experimental metadata, initial/target conditions, operational procedures, state machine evolution, process data, chemical quantity records, key event logs, anomaly records, experimental results (final pH value, total volume, pKa value, deviation), and reproduction guidelines.

\textbf{Step 3 - Output:} Generate a structured report (JSON/Markdown format) and clearly label any anomaly information.

Ensure the report is complete and supports full experiment reproduction.''
\end{quote}

\subsubsection{Action Agent Training Method}

The VLA model of the Action Agent was pre-trained on a computing cluster consisting of eight NVIDIA H20 GPUs, using a global batch size of 1024 and performing 200,000 iterations with bfloat16 mixed precision. The training took approximately four days. The model was optimized using the AdamW optimizer (\(\beta_1 = 0.9, \beta_2 = 0.95\)), with a learning rate set to \(1 \times 10^{-4}\) and weight decay of 0.01. All input images were resized to a resolution of \(224 \times 224\), and lightweight ColorJitter data augmentation was applied to improve generalization robustness.

During the pre-training phase, the Action Agent used a carefully constructed heterogeneous data mix, consisting of a total of 290,000 robot operation trajectories. These data were sourced from the Droid, Robomind, and Agibot large-scale robot datasets, covering 7 robot platforms and 5 robotic arm configurations, including single-arm and dual-arm collaboration systems. By introducing soft prompts on a unified backbone network to absorb differences across hardware forms, the model learned a general operational strategy with cross-form generalization capabilities, laying the foundation for subsequent domain adaptation and deployment.

\section{Conclusion}\label{sec13}

In this work, we identified the long-tail challenge in laboratory automation, characterized by limitations at the task, chemical perception, and environmental levels. To address these, we introduced AgentChemist, a multi-agent robotic platform that integrates planning, real-time monitoring, precision control, and automated reporting. By treating chemical experiments as finite state machines, AgentChemist dynamically decomposes tasks, adapts to evolving reaction states, and operates across diverse laboratory setups without specialized infrastructure. Validation through titration and solid weighing experiments demonstrated autonomous progress tracking, adaptive control, and reliable execution of multi-step procedures, with continuous operation for up to eight hours without human intervention. These results highlight the potential of multi-agent systems to overcome the rigidity of traditional automation, enabling more flexible, perceptive, and resilient chemical experimentation. AgentChemist thus represents a step toward intelligent, collaborative laboratory automation capable of handling the variability and complexity inherent in real-world research.

\backmatter

\bibliography{sn-bibliography}

\end{document}